\PassOptionsToPackage{dvipsnames}{xcolor}
\PassOptionsToPackage{inline}{enumitem}
\documentclass{article}

\usepackage[eandd, final]{neurips_2026}

\usepackage[utf8]{inputenc} 
\usepackage[T1]{fontenc}    
\usepackage{hyperref}       
\usepackage{url}            
\usepackage{booktabs}       
\usepackage{tabularx}
\usepackage{wrapfig}
\usepackage{amsfonts}       
\usepackage{nicefrac}       
\usepackage{microtype}      
\usepackage{xcolor}         
\usepackage{soul}
\usepackage{url}
\usepackage[utf8]{inputenc}
\usepackage{xspace}
\usepackage{graphicx}
\usepackage{amsmath}
\usepackage{amsthm}
\usepackage{amssymb}
\usepackage{booktabs}
\usepackage{algorithm}
\usepackage{algorithmic}
\usepackage{xcolor}
\usepackage{listings}
\usepackage{float}
\usepackage{fontawesome5}
\usepackage{colortbl}
\usepackage{pgffor}
\usepackage{enumitem}
\usepackage{multirow}
\usepackage{arydshln}
\usepackage{listings}
\usepackage{xcolor}

\newif\ifrevise
\revisefalse   

\newcommand{\revision}[1]{%
  \ifrevise
    \textcolor{red}{#1}%
  \else
    #1%
  \fi
}

\usepackage{pifont}
\newcommand{\cmark}{\textcolor{black}{\ding{51}}} 
\newcommand{\xmark}{\textcolor{black}{\ding{55}}} 

\usepackage{tikz}

\usepackage{circledsteps}

\newcommand{\spara}[1]{\vspace{0.1cm}\noindent{\bf #1}}

\definecolor{aliceblue}{rgb}{0.94, 0.97, 1.0} 
\definecolor{azure(colorwheel)}{rgb}{0.0, 0.5, 1.0} 
\definecolor{aureolin}{rgb}{0.99, 0.93, 0.0} 

\lstdefinelanguage{json}{
    basicstyle=\ttfamily\scriptsize, 
    numbers=left, 
    numberstyle=\tiny\color{gray}, 
    numbersep=-8pt, 
    xleftmargin=0em, 
    columns=flexible, 
    breaklines=true,
    frame=single,
    backgroundcolor=\color{aureolin!4},
    literate=
      *{"class"}{{\textcolor{blue}{"class"}}}{1}
      {"parameters"}{{\textcolor{teal}{"parameters"}}}{1}
      {"partitions"}{{\textcolor{violet}{"partitions"}}}{1}
      {"data"}{{\textcolor{red}{"data"}}}{1}
      {"unlearners"}{{\textcolor{red}{"unlearners"}}}{1}
      {"predictor"}{{\textcolor{red}{"predictor"}}}{1}
      {"evaluator"}{{\textcolor{red}{"evaluator"}}}{1}
}

\lstdefinelanguage{myPython}[]{Python}{
    basicstyle=\ttfamily\scriptsize, 
    numbers=left, 
    numberstyle=\tiny\color{gray}, 
    xleftmargin=0em,
    breaklines=true,
    frame=single,
    backgroundcolor=\color{aureolin!4},
    stringstyle=\color{purple},
    commentstyle=\color{teal}
}
\usepackage{mdframed}
\usepackage{amssymb}
\usepackage{comment}
\usepackage{wrapfig}

\newtheorem{definition}{Definition}

\newcommand{\our}[0]{Lethe\xspace}
\newcommand{\gold}{M_G}
\newcommand{\unlearned}{M_u}
\newcommand{\model}{M^*}
\newcommand{\forget}{E_f}

\newcommand{\numberofmethods}[0]{15 }
\newcommand{\numberofdatasets}[0]{8 }
\newcommand{\numberofarchitectures}[0]{5 }

\mdfdefinestyle{stebox1}{%
    linecolor=black!70,
    linewidth=1.2pt,
    leftmargin=0cm,
    rightmargin=0cm,
    roundcorner=2pt,
    innerleftmargin=10pt,
    innerrightmargin=10pt,
    innertopmargin=6pt,
    innerbottommargin=6pt,
    topline=false,
    bottomline=false,
    rightline=true,
    backgroundcolor=gray!8
}

\usepackage[most]{tcolorbox}

\newtcolorbox{rqbox1}{
  colback=white,
  colframe=black,
  left=4pt,
  right=4pt,
  top=4pt,
  bottom=4pt,
  boxrule=0pt,
  borderline west={2pt}{0pt}{black}
}

\newcommand{\rqbox}[2]
{   \vspace{5pt}
    \begin{mdframed}[style=stebox1]
    {\textit{#1}}
    \end{mdframed}
    \vspace{5pt}
}

\title{\our: Link Inference Attacks For Evaluation of Edge Unlearning Methods}

\author{%
  Andrea D'Angelo\\
  University of Aarhus\\
  Aarhus, Denmark\\
  \texttt{andrea@cs.au.dk} \\
  \And
  Francesco Gullo\\
  University of L'Aquila\\
  L'Aquila, Italy\\
  \texttt{francesco.gullo@univaq.it} \\
  \And 
  Davide Mottin\\
  University of Aarhus\\
  Aarhus, Denmark\\
  \texttt{davide@cs.au.dk} \\
  \And
  Sayan Ranu\\
  Indian Institute of Technology Delhi\\
  Delhi, India\\
  \texttt{sayanranu@iitd.ac.in} \\
  \And
  Giovanni Stilo\\
  Luiss Guido Carli University\\
  Rome, Italy\\
  \texttt{gstilo@luiss.it}
}

\begin{document}

\maketitle

\begin{abstract}
Graph Unlearning (GU) aims to remove the influence of specific training data from a Graph Neural Network without retraining anew. While most methods address removal of nodes and features, Edge Unlearning (EU), aiming at forgetting edges, has recently attracted dedicated attention as a standalone problem due to the hardness of completely removing the impact of an edge in the model. Despite this, no dedicated benchmark for EU exists besides isolated evaluations on feature-rich datasets such as Cora and Citeseer where GNNs exploit node features rather than graph structure. As such, we show two systematic limitations in existing evaluations: dedicated EU methods can be slower than retraining anew, and accuracy metrics are uninformative due to the negligible impact of edge removal. To address this gap, we introduce \our\footnote{River of forgetfulness in Greek mythology.}, the first comprehensive benchmark for Edge Unlearning, featuring \numberofmethods methods and  \numberofdatasets datasets. Our benchmark emphasizes large graphs, introduces evaluation tasks of varying difficulty, and proposes empirical measures based on link inference attacks rather than proxy accuracy metrics. We provide a comprehensive empirical analysis, demonstrate the shortcomings of current evaluations, and release \our as a reproducible, extensible benchmark.
\end{abstract}

\section{Introduction}

Graph Unlearning (GU) methods aim to remove the influence of a subset of training data from a Graph Neural Network (GNN) without retraining it from scratch. This problem has gained increasing popularity due to recent regulatory requirements such as the European General Data Protection Regulation (GDPR), which enforces the \textit{right to be forgotten}, allowing users to request the removal of their data from trained models \cite{gdpr_problems}.

GU has attracted significant attention \cite{graph_unlearning} due to the widespread use of graph-structured data in applications involving sensitive information, such as social networks \cite{social_networks} and medical records \cite{medical_networks}. In such scenarios, the scale of data makes retraining a GNN from scratch prohibitively expensive in terms of time and computational resources \cite{env_costs}. GU methods have emerged as the prime alternative, enabling the removal of specific nodes or edges without full retraining.

In this paper, we focus on \textbf{Edge Unlearning} (EU) as a fundamental subproblem of GU that targets the removal of information associated with specific edges in the graph. In many applications, sensitive information is inherently relational and encoded in edges rather than individual nodes.
For instance, in a social network, a user may request the removal of a connection (e.g., a past interaction). In this case, the model holder must eliminate the influence of that specific edge while preserving the remaining graph structure \cite{ceu,IDEA}.
While most GU methods address both node and edge unlearning, the latter has recently attracted dedicated attention due to the inherent hardness of ensuring that the removed edge cannot be inferred through other mechanisms, with several methods proposed exclusively for this setting \cite{unlink_to_unlearn,ceu}.

Despite its importance, there is currently no dedicated benchmark for EU.  Existing EU methods construct the experimental setups without standardized protocols, often relying on widely used datasets that are neither consistently adopted across studies nor rigorously evaluated for this task. The most recent benchmark by Jain et al. \cite{gu_benchmark} focuses  on \emph{node unlearning}; this work offers a complementary view by providing the set of necessary tools for evaluating GU on \textbf{both} node and edge unlearning. While \cite{open_gu} includes both node and edge unlearning, its evaluation relies primarily on UMIA and accuracy metrics, limiting the breadth of its evaluation protocol and motivating the shortcomings discussed below.

\begin{figure}[t]
    \centering    \includegraphics[width=0.8\linewidth]{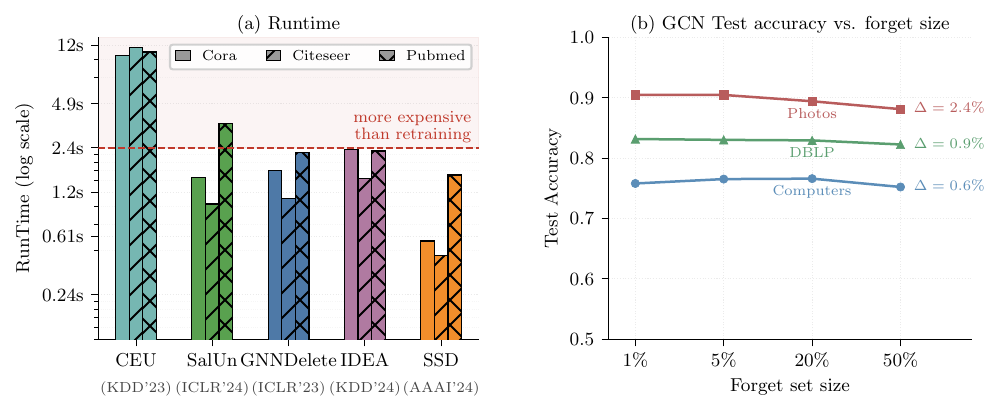}
    \caption{Main benchmark findings. (a) Inefficiency: Some methods take longer than retraining the model from scratch (Gold Model) or exhibit negligible runtime advantage; (b) Uninformative measures: No significant drop in accuracy after removing up to 50\% of edges.}
    \label{fig:intro_figure}
\end{figure}

Critically, we find that the evaluations in the literature suffer from inherent limitations. Unlike node or feature unlearning, EU is fundamentally structure-driven: removing an edge alters the graph topology, leaving all nodes and node features intact. As a result, the effectiveness of an unlearning method depends on whether the model actually relies on structural (i.e., edges, subgraphs) signals. Existing works predominantly evaluate on small, feature-rich datasets like Cora \cite{Cora}, Citeseer \cite{Citeseer} and Pubmed \cite{pubmed}, where GNNs can effectively leverage node features rather than structural information, rendering edge removal largely inconsequential. This leads to two major problems, which we visualize in Figure \ref{fig:intro_figure}: \textbf{(a)} The runtime of methods on these datasets is comparable (and sometimes higher) to that of the \emph{Gold Model} obtained retraining the model anew, defeating the purpose of using unlearning methods. The only marginal gains are typically below one second and thus not practically meaningful. \textbf{(b)} On bigger datasets, the impact of edge removal on accuracy is negligible, even retraining from scratch on a graph with up to half of the edges removed (which is unrealistic) results in stable accuracy on most datasets ($\Delta \leq 5\%$). This is consistent with the literature on GCNs \cite{gcn}, notably dropping edges for boosting performance (like in the case of DropEdge \cite{dropedge}), and shows that accuracy reported by existing GU methods on such datasets is not informative for evaluating unlearning effectiveness.
These observations suggest that the settings used to evaluate EU methods are not fully aligned with the EU task, leading to potentially fallacious conclusions about method's effectiveness and efficiency.


To fill this gap, we present \our, the first dedicated benchmark for EU, spanning \numberofmethods methods, \numberofdatasets datasets, and \numberofarchitectures model architectures. 
\our\ solves both problems by \textbf{(i)} only employing datasets big enough for runtime to matter, \textbf{(ii)} employing three link inference attacks that are more informative than accuracy alone on most datasets and \textbf{(iii)} proposing different forget set selection strategies based on an edge centrality measure called WalkCentrality.
\our\ is openly available\footnote{\url{https://github.com/dangeloandrea14/Lethe}}, reproducible, extensible and focuses on structural properties of graphs rather than feature-rich representations. \our\ sheds light on the efficacy of existing MU and GU methods, surpassing standard, accuracy-based evaluations on small graphs. Specifically, we show how link inference attacks highlight several criticalities across methods, and how selecting the forget set based on a centrality measure can lead to systematic failures in defending from such attacks. 

\section{Related Work}

\spara{Edge Unlearning.}
Graph Unlearning methods aim to approximate the effect of retraining a GNN from scratch after removing a subset of training data \cite{graph_unlearning}.
Existing approaches for GU that are also defined and evaluated on EU either prove their effectiveness empirically, like in the case of GNNDelete \cite{GNNDelete}, or provide theoretical certification of the closeness of their solution to the Gold Model, like in the case of CGU \cite{cgu}, IDEA \cite{IDEA}, and ScaleGUN \cite{scalegun}.

While many GU methods handle edge unlearning as one supported request type, a dedicated line of work targets it as a standalone problem. 
GNNDelete~\cite{GNNDelete} formalizes two structural objectives: \textit{deleted edge consistency}, requiring that predictions for removed edges resemble those for non-existent edges, and \textit{neighborhood influence}, ensuring predictions in the local vicinity of a removed edge remain stable.
CEU~\cite{ceu} provides certified guarantees for edge unlearning via quasi-Newton parameter updates with Gaussian noise injection to mask gradient residuals.
Unlink to Unlearn~\cite{unlink_to_unlearn} avoids parameter modification entirely, instead masking the forgotten edge during message passing at inference time. 

All of these methods, both for GU and EU, employ small, feature-rich datasets like Cora \cite{Cora} and Citeseer \cite{Citeseer} in their experimental evaluation, and argue that the results generalize to large-scale graphs. ScaleGUN is the only one observing how these small datasets are not enough for a fair evaluation. We bring this observation to its logical conclusion: the community needs a benchmark for evaluating EU methods in fair condition on these datasets. Moreover, several general-purpose Machine Unlearning algorithms are applicable to GNNs. These methods include \textbf{(i).} fine-tuning the GNN on the retain graph or, conversely, applying Gradient Ascent on the forget set of the graph \cite{finetune}, \textbf{(ii).} modifying the weights of the GNN directly, like in the case of SalUn \cite{salun} and UNSIR \cite{unsir}, \textbf{(iii).} distillation-based methods with a teacher model, e.g. Bad Teaching \cite{badteaching} and Scrub \cite{scrub}. Ours is the first work to benchmark them for edge removal and compare them to dedicated GU methods.

\spara{Benchmarking of Graph Unlearning.} Despite increasing attention on the topic, GU (and EU consequently) has received limited surveying and benchmarking efforts. 
\cite{gu_benchmark} proposes the most recent GU benchmark concluding that most GU methods are not ready for practical deployment yet, due to poor scaling on large graphs. However, their work focuses only on the task of node unlearning while EU is essential for a complete evaluation of graph unlearning methods, as it addresses cases that node unlearning cannot model. Our study complements the existing benchmark \cite{gu_benchmark} by defining the protocol for the edge setting, thereby covering an important class of scenarios that node unlearning alone cannot capture, in which graph size and edge density are even more critical. We aim to provide the community with all the necessary tools to fully evaluate their GU methods, both on node and edge unlearning. Similarly, OpenGU~\cite{open_gu} provides a general benchmark for both tasks. However, while OpenGU focuses on the membership inference attack (MIA) from general machine unlearning, our analysis suggests that MIA is often the least informative in a graph context. We address this by implementing three dedicated link inference attacks specifically designed for the graph unlearning setting, and two strategies for centrality-guided forget set selection. Consequently, we provide all methods and attacks within a standardized, reproducible, and extensible framework. 

\spara{Link Inference Attacks.}
Unlearning efficacy in the edge setting is most directly assessed via link inference: adversarial probing of a model's outputs to determine whether a specific edge was present during training.
Prior work has shown that GNNs are susceptible to such attacks through shadow model approaches~\cite{stealing_links}, influence function-based methods~\cite{linkteller}, and general inference frameworks~\cite{linkinferenceattacks}.
In EU evaluation, link inference attacks serve as the adversarial counterpart to accuracy-based utility metrics, providing a direct test of whether edge information has been forgotten, or if it can be adversarially reconstructed.
However, existing works adopt incompatible threat models and attack implementations, making cross-method comparison difficult.
\our\ aims to standardize this evaluation component to guide future research in the field.

\spara{Critique of current benchmarks.} Many prior works in the literature of GNNs highlight the increasing necessity of evolving the evaluation techniques in the field. Palowitch et al. \cite{graphworld} argue that GNNs are usually evaluated on the same handful of small datasets with low structural information. Coupette et al. \cite{nometrictorulethemall} stress the importance of \textit{good} graph benchmarks, with a focus on structural signals, something that the most popular datasets are still missing. Bechler-Speicher et al. \cite{graphlearningwillloserelevance}, in their position paper, argue that graph learning is at risk of losing relevance because of how poorly designed graph learning benchmarks are, the over-reliance on the same few datasets, and the neglect of structurally interesting graphs. This need naturally carries over to the task of EU, where forgetting means being unable to reconstruct the edge from any unremoved graph signal. As opposed to previous benchmarks focusing exclusively on node unlearning, our focus is to collect a set of datasets designed for profiling EU, with the aim of complementing GU with measures based on link inference attacks and dataset hardness measures to help the community quantify progress in EU in a systematic manner.



\section{Limitations of Current Evaluations}\label{sec:flaws_current_evaluations}

In this section, we provide empirical evidence of two main limitations in the evaluations of EU in the literature, namely the usage of graphs that are either too small and/or over-reliant on node features.  
Due to the lack of benchmarks for EU, most works define their own experimental settings, usually relying on popular datasets that are often used in the context of graph learning. Indeed, most edge unlearning methods evaluate on standard benchmarks like Cora, Citeseer, and Pubmed~\cite{graph_unlearning}, relying on the test accuracy of the unlearned model as a proxy for efficacy~\cite{ceu,cgu,IDEA,GNNDelete}.  However, these evaluation settings are not well aligned with the EU goals as they falter in one of two distinct ways:

\paragraph{(i) Small-scale datasets: negligible efficiency gains.}
Datasets such as Cora, Citeseer and Pubmed are extremely small, making retraining from scratch very cheap. Even in the original papers, reported training times for datasets such as Cora and Citeseer typically range from $3$ to $5$~seconds \cite{ceu}.

\begin{figure}
    \centering
    \includegraphics[width=\linewidth]{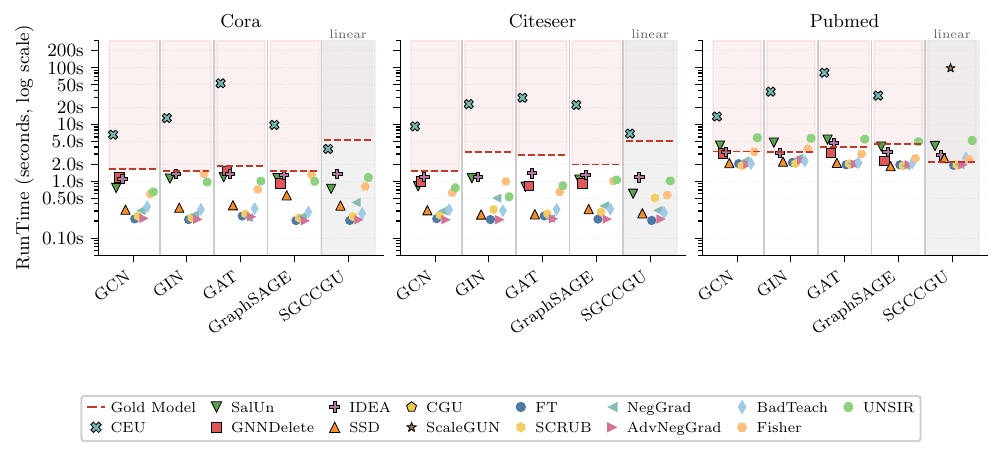}
    \caption{Runtime (sec.) of GU methods on Cora, Citeseer and Pubmed across 5 different GNN architectures. Note that CGU and ScaleGUN only work on linear models (i.e., SGC). The red line represents the Gold Model: anything above implies that it takes more time than retraining the model from scratch.}
    \label{fig:runtimes}
\end{figure}

In our empirical evaluation (Figure \ref{fig:runtimes}) most methods, both general MU and GU methods, run in time comparable to that of retraining on Cora, Citeseer, and Pubmed, with simpler methods like Fine-tuning being slightly faster. However, such gains are not only negligible in practice, but also unlikely to generalize. Indeed, differences of this magnitude can be attributed to statistical noise or machine- and implementation-specific factors, such as hardware variability or framework overhead (e.g., PyTorch initialization latency), rather than meaningful algorithmic efficiency. The graph density of these datasets also plays a role in the runtime overhead. CEU~\cite{ceu} introduces the notion of \textit{infected nodes} in their implementation, as the nodes affected by the removal of a forgotten edge within $k$-hops, where $k$ is the number of GNN layers. However, removing even 1\% of edges in these high-density graphs infects almost all nodes within two hops, meaning that MU methods are effectively operating on almost the entire graph, leaving little room to amortise the cost of unlearning against the cost of retraining. We show this in detail in Section \ref{apx:infected}. 

\begin{table}[t]
\centering
\caption{Gold Model test accuracy (\%) across architectures and datasets
         at 5\%, 20\%, and 50\% forget-set size.
         Original shows the accuracy of the unmodified model (upper bound).
         $\Delta$ is the accuracy change relative to Original.}
\label{tab:goldmodel_accuracy}
\scriptsize
\setlength{\tabcolsep}{5pt}
\begin{tabular}{@{}ll r rr rr rr@{}}
\toprule
\multirow{2}{*}{Dataset} & \multirow{2}{*}{Architecture} & \multirow{2}{*}{Original} & \multicolumn{2}{c}{Gold Model (5\%)} & \multicolumn{2}{c}{Gold Model (20\%)} & \multicolumn{2}{c}{Gold Model (50\%)} \\
\cmidrule(lr){4-5}\cmidrule(lr){6-7}\cmidrule(lr){8-9}
 & & & Acc & $\Delta$ & Acc & $\Delta$ & Acc & $\Delta$ \\
\midrule
\multirow{5}{*}{Computers} & GCN & 79.7 & 81.1 & +1.3 & 81.6 & +1.9 & 75.2 & -4.5 \\  
 & GIN & 42.7 & 43.8 & +1.1 & 25.0 & -17.6 & 36.6 & -6.1 \\
 & GAT & 81.8 & 81.4 & -0.4 & 80.9 & -0.9 & 79.1 & -2.7 \\
 & GraphSAGE & 84.0 & 83.9 & -0.1 & 83.4 & -0.5 & 81.1 & -2.9 \\
 & SGCCGU & 75.6 & 74.4 & -1.2 & 74.6 & -1.1 & 73.5 & -2.2 \\
\midrule
\multirow{5}{*}{Photos} & GCN & 91.0 & 89.1 & -2.0 & 90.3 & -0.8 & 88.1 & -2.9 \\
 & GIN & 66.3 & 74.5 & +8.2 & 70.7 & +4.4 & 65.2 & -1.0 \\
 & GAT & 91.6 & 90.9 & -0.7 & 90.4 & -1.2 & 88.9 & -2.7 \\
 & GraphSAGE & 93.5 & 92.9 & -0.7 & 92.2 & -1.4 & 90.3 & -3.2 \\
 & SGCCGU & 87.3 & 87.4 & +0.1 & 86.3 & -1.0 & 85.1 & -2.2 \\
\midrule
\multirow{5}{*}{DBLP} & GCN & 82.5 & 82.6 & +0.1 & 83.1 & +0.6 & 82.3 & -0.2 \\
 & GIN & 83.7 & 83.2 & -0.5 & 83.9 & +0.2 & 82.4 & -1.3 \\
 & GAT & 83.4 & 82.3 & -1.2 & 82.3 & -1.2 & 81.9 & -1.5 \\
 & GraphSAGE & 84.3 & 84.4 & +0.1 & 84.3 & +0.0 & 83.9 & -0.4 \\
 & SGCCGU & 71.0 & 70.6 & -0.4 & 70.3 & -0.6 & 69.7 & -1.2 \\
\bottomrule
\end{tabular}
\end{table}

\paragraph{(ii) Feature-dominated datasets: negligible impact on accuracy.}
A second, orthogonal issue arises in datasets characterized by highly informative node features, typically in the form of bag-of-words representations or word embeddings. In such settings, node features alone are often sufficient to achieve strong predictive performance, reducing the relative importance of graph structure. 

Table \ref{tab:goldmodel_accuracy} shows the Gold Model test accuracy across all architectures and datasets when the forget set is 5\% and 20\% of the training set. 
For most architectures the change in accuracy is within 3\%, except for GIN \cite{gin} which does not normalize across neighbor embeddings and is thus more affected by edge removals.
In general, removing edges has only a limited effect on model predictions, leading to negligible changes in accuracy. This makes it difficult to meaningfully evaluate the effectiveness of EU methods through this proxy metric, as the task itself becomes insensitive to structural modifications, so relying on accuracy only can be misleading.

\section{\our}

Our benchmarking methodology comprises two sections. Section~\ref{sec:design} elaborates on the observations in Section \ref{sec:flaws_current_evaluations}  and introduces \our, our benchmark that fills the literature gap (refer to Section \ref{apx:attacks} for a detailed description of the link inference attacks included). Section \ref{sec:difficulty} describes the two forget set selection strategies included in \our.

\subsection{Design of \our}\label{sec:design}

\begin{table}[t]
\centering
\scriptsize
\caption{%
  Datasets included in \our\ and their key properties.
  \textit{Avg.\ deg}: average node degree.
  \textit{Density}: edge density.
  $\Delta_{20\%}$: accuracy change Gold$-$Identity (pp) at $k{=}20\%$ forget set.
}
\label{tab:datasets}
\setlength{\tabcolsep}{5pt}
\begin{tabular}{l l r r r r r r r}
\toprule
Dataset & Domain
  & $|V|$ & $|E|$ & Avg.\ deg & Density & Features & Classes
  & $\Delta_{20\%}$ \\
\midrule
Cora
  & Citation
  & 2.7K & 10.6K & 7.8 & 0.29\% & 1433 & 7
  & $-3.1$ \\
Citeseer
  & Citation
  & 3.3K & 9.2K & 5.5 & 0.17\% & 3703 & 6
  & $-1.5$ \\
Pubmed
  & Citation
  & 19.7K & 44.3K & 4.5 & 0.02\% & 500 & 3
  & $+0.3$ \\
\midrule
Computers
  & Co-purchase
  & 13.8K & 492K & 35.8 & 0.52\% & 767 & 10
  & $+0.5$ \\
Photos
  & Co-purchase
  & 7.6K & 238K & 31.1 & 0.81\% & 745 & 8
  & $-2.2$ \\
DBLP
  & Citation
  & 17.7K & 106K & 6.0 & 0.07\% & 1639 & 4
  & $+0.1$ \\
Flickr
  & Social
  & 89K  & 900K & 10.1 & 0.02\% & 500 & 7
  & $-2.3$ \\
arxiv
  & Citation
  & 169K & 1.2M & 7.1 & 0.01\% & 128 & 40
  & $-0.6$ \\
\bottomrule
\end{tabular}
\end{table}

The findings of Section \ref{sec:flaws_current_evaluations} highlight a set of concrete requirements
that a benchmark for Edge Unlearning must satisfy. We state the three following desiderata and use them to construct \our.

\paragraph{$D_1$: Scale and graph density.}
The practical relevance of MU depends on retraining being expensive. On small
graphs, as shown in Figure \ref{fig:runtimes}, retraining from scratch
typically takes only a few seconds, making any unlearning overhead negligible in
absolute terms. \our\ prioritises graphs of sufficient size (>5K nodes, >100K edges) in which the Gold Model
constitutes a genuine computational cost. For this reason, while \our\ supports Cora, Citeseer, and Pubmed, we omit them from the discussion of results. Table~\ref{tab:datasets} summarises the datasets included in \our\ and their alignment with $D_1$.

\paragraph{$D_2$: Structural informativeness.}
On feature-rich datasets, node embeddings alone carry sufficient discriminative signal, and the graph structure acts as a weak regularizer. Edge removal then has
negligible effect on the model, making accuracy metrics
uninformative. 
Crucially, EU is inherently a \emph{privacy} problem. The objective is not only to maintain performance on downstream tasks, but to ensure that the model no longer encodes information about the forgotten edges. Accuracy alone cannot capture this notion of forgetting, as it does not quantify whether structural information can still be inferred from the model.
To address this gap, we adapt three established link inference attacks to measure whether a method is effectively unlearning: UMIA \cite{scrub}, Link  Stealing \textbf{(LS)} \cite{stealing_links}, and LinkTeller \textbf{(LT)} \cite{linkteller}. We assume access to the unlearned model and retain graph, and the adversary must classify a candidate edge $(u, v)$ as either a forgotten edge (positive) or a non-existing edge (negative). Details and implementation of these attacks are available in Appendix \ref{apx:attacks}. We report the ROC-AUC of these scores over positive (forgotten) and negative (non-existing) pairs.

\paragraph{$D_3$: Task difficulty diversity.}
Not all edges are equally hard to unlearn. \revision{Edges that appear in sparser zones and are not structurally redundant} have
a \revision{strong} footprint in the learned representations and are \revision{less} straightforward
to forget. Conversely, highly-central edges participate in many message-passing
paths and leave a \revision{weaker} trace across node representations. A rigorous
benchmark must evaluate methods across this spectrum. To this end, we define the metric of WalkCentrality (Section \ref{sec:difficulty}) as a yardstick to implement two forget set selections strategies in \our.

Following $D_1-D_3$, \our\ includes \numberofdatasets datasets and \numberofmethods methods, assessed using three complementary link inference attacks, each evaluated under two different settings (low and high centrality, explained in Section \ref{sec:difficulty}).

\subsection{Settings of Edge Unlearning: WalkCentrality}\label{sec:difficulty}

Current works simply sample a random percentage (usually either 5\% or 20\%) of edges to build the forget set. However, not all edges are equally important, given the message passing systems of GNNs. \revision{Intuitively, some edges will participate in more message passing paths than others. To model this property in \our, we introduce two different difficulty settings for edge unlearning, leveraging the model-agnostic measure of WalkCentrality.}

\revision{We ask: \textit{given an edge e, how redundant is it in the message passing mechanism?} This gives us an intuition of how much ``information'' the edge will carry during the message passing, and how much its propagation is smoothed from redundant paths. }

Since GCNs propagate features using the symmetrically normalized adjacency matrix $\hat{A} = \tilde{D}^{-1/2}\tilde{A}\tilde{D}^{-1/2}$, where $\tilde{A} = A + I$ adds self-loops, we use $\hat{A}$ in place of the raw adjacency matrix. This captures both degree normalization and self-loops, giving a more faithful measure of how much a given edge influences the learned representations. Any $k$-step walk passing through edge $(i \to j)$ at step $t$ decomposes uniquely into a $t$-step walk from some source to $i$, the edge $(i \to j)$, and a $(k-t-1)$-step walk from $j$ to some destination. \revision{Summing the weights of these walks over all $t$, excluding the factor $\hat A_{ij}$ of the edge itself (so that the score measures the propagation surrounding the edge), gives the walk-based centrality for edge (i, j) across a k-layer GCN:} 








\[
\text{WalkCentrality}(i,j) = \sum_{t=0}^{k-1} \left(\mathbf{1}^\top \hat{A}^t\right)_i \cdot \left(\hat{A}^{k-t-1} \mathbf{1}\right)_j
\]

This is efficient to compute: precomputing $\{\hat{A}^0\mathbf{1}, \ldots, \hat{A}^{k-1}\mathbf{1}\}$ and $\{\mathbf{1}^\top \hat{A}^0, \ldots, \mathbf{1}^\top \hat{A}^{k-1}\}$ reduces the centrality of every edge to a simple lookup and sum, requiring $O(k \cdot |E|)$ total for sparse graphs.
WalkCentrality allows us to score and rank all edges in a graph. By selecting 5\% of edges with the lowest/highest WalkCentrality, we configure the \textit{low/high} centrality settings of \our.

\revision{\paragraph{Observation.} Ignoring learnable weights for simplicity, a $k$-layer GNN with normalized adjacency message passing computes: $H^{(k)}= \hat{A}^kX$, where $\hat{A}$
is the normalized adjacency matrix and the entry $\hat{A}^k_{uv}$
is the aggregate contribution of all length-$k$ walks from $u$ to $v$. For an edge $(i,j)$, the term $(\mathbf{1}^{\top}\hat{A}^{t})_i$
is how much signal reaches $i$ after $t$ steps of propagation, and
$(\hat{A}^{k-t-1}\mathbf{1})_j$ is how much signal spreads out of in the steps that remain. Summing their product over every way of splitting the steps between the two endpoints gives}
$\sum_{t=0}^{k-1}(\mathbf{1}^{\top}\hat{A}^{t})_i(\hat{A}^{k-t-1}\mathbf{1})_j,$
\revision{which is exactly the WalkCentrality score used in \our.}

\revision{Because $(\hat{A}^{t}\mathbf{1})_i$
grows with the connectivity of $i$ and of its $t$-hop neighbourhood, a high score means the edge sits between endpoints that are already carrying a large volume of propagated signal and therefore that its own share of that signal is small. This Observation shows how WalkCentrality is directly derived from a mechanistic interpretation of the GNN's message passing system.}


\revision{To measure this directly rather than only through link inference attacks, we ran experiments on three datasets (Photos, Computers and DBLP, with a GCN, 5\% forget set), showing that WalkCentrality is strongly negatively correlated with training loss, and that low-centrality removals are localized and have a large impact. Results are reported in Appendix \ref{apx:walkcentrality}.}

\section{Experimental Evaluation}\label{sec:experiments}

We conduct the experiments on a Linux-based high-performance computing cluster equipped with Intel Xeon Gold 6140M CPUs (2.30 GHz, 108 cores), 880 GB of RAM, and an NVIDIA A100 GPU (80 GB memory). We tune the hyperparameters on the learning rate of general MU methods to adapt them to GNNs. 
For graph-specific methods, we use the settings described in the original papers. \revision{A broad overview of the methods that we use is present in Appendix \ref{apx:methods}}. 
All the architectures are trained for 100 epochs with the Adam optimizer with early stopping using a patience of 10 epochs and a minimum improvement threshold of 0.01. 
Original and Gold Models are cached between methods and settings to provide a stable reference. 

\revision{Specifically, the Gold Model and attack-based metrics evaluate two complementary but distinct properties of an unlearning method (refer to the definition of EU in Appendix \ref{apx:definitions}). The Gold Model represents the outcome of retraining from scratch after removing the forgotten edges and therefore serves as the reference for retraining fidelity. Attack-based metrics, instead, ask: \textit{after unlearning, can an adversary still recover information about the edges that were requested to be forgotten?}}

\revision{Therefore, the Gold Model provides the reference for functional behavior after forgetting, whereas attack-based metrics measure privacy leakage (the ideal value being $0.5$, so the attack is basically random guessing). Importantly, this is also the case for the most recent node unlearning benchmark of \cite{gu_benchmark}, that we aim to complement: they also consider an attack AUC of 0.5 as ideal, regardless of the Gold Model. We adapt to this standard, which is consistent with the unlearning literature}.

The methods we benchmark in this experimental evaluation are either general-purpose MU methods and GU methods. 
As general-purpose MU methods, we evaluate Fine-tuning \cite{finetune}, Negative Gradient \cite{finetune}, Advanced Negative Gradient \cite{finetune}, UNSIR \cite{unsir}, BadTeaching \cite{badteaching}, SCRUB \cite{scrub}, Fisher Forgetting \cite{ssd}, Selective Synaptic Dampening \cite{ssd}, SalUn \cite{salun}. 
For GU, we evaluate IDEA \cite{IDEA}, CGU \cite{cgu}, CEU \cite{ceu}, ScaleGUN \cite{scalegun}, UnlinkToUnlearn \cite{unlink_to_unlearn} and GNNDelete \cite{GNNDelete}. The application of general MU methods to Edge Unlearning is a unique contribution of this work. 

\revision{In Appendix \ref{apx:ablations}, we report the results of all the ablation studies we perform, and their consequences on the Research Questions we answer in this section. Specifically, we ablate: \textbf{(i)}. The attacker's accessible information for link inference attacks in Appendix \ref{apx:attacker_ablation}, \textbf{(ii).} The forget set selection strategy, in Appendix \ref{apx:forget_strategy}, and \textbf{(iii).} The downstream task, in Appendix \ref{apx:link_prediction}.}


\subsection{Results}\label{sec:results}

Here we outline the findings of the experiments using \our. We report the results obtained on GCNs, which is the most popular GNN architecture, with a forget set of 5\%. We aim to answer three research questions targeting fundamental open issues in EU evaluation: 

\begin{itemize}[noitemsep,topsep=0pt,leftmargin=*]
\item \textbf{RQ1}: Are link inference attacks more informative than accuracy? Which one is the most effective?
\item \textbf{RQ2}: Are the task centrality settings informative?
Do WalkCentrality-based settings reveal qualitatively different behaviours?
\item \textbf{RQ3}: Are certified EU methods applicable in practice and resilient to link inference attacks?
\end{itemize}







\begin{table*}[t]
\centering
\footnotesize
\caption{%
  Accuracy and link inference attacks across all methods.
  Upper block: GCN architecture. Lower block: SGC architecture.
  Acc: test accuracy (\%\,; $\uparrow$ better).
  UMIA: E-UMIA AUC ($\approx$0.5 = indistinguishable from random).
  LT: LinkTeller AUC.
  LS: Link Stealing AUC.
  \colorbox{gray!15}{Shaded rows} are reference baselines. $^{\S}$ GNNDelete collapses model performance, as found in \cite{gu_benchmark}.
}
\label{tab:rq1}
\scriptsize
\setlength{\tabcolsep}{2pt}
\resizebox{\textwidth}{!}{%
\begin{tabular}{l cccc cccc cccc cccc cccc}
\toprule
& \multicolumn{4}{c}{Photos}
& \multicolumn{4}{c}{Computers}
& \multicolumn{4}{c}{DBLP}
& \multicolumn{4}{c}{Flickr}
& \multicolumn{4}{c}{ogbn-arxiv} \\
\cmidrule(lr){2-5}\cmidrule(lr){6-9}\cmidrule(lr){10-13}\cmidrule(lr){14-17}\cmidrule(lr){18-21}
Method
  & Acc & UMIA & LT & LS
  & Acc & UMIA & LT & LS
  & Acc & UMIA & LT & LS
  & Acc & UMIA & LT & LS
  & Acc & UMIA & LT & LS \\
\midrule
\rowcolor{gray!15}
Original
  & 90.8 & 0.526 & 0.480 & 0.639
  & 79.9 & 0.500 & 0.564 & 0.525
  & 82.5 & 0.505 & 0.392 & 0.733
  & 49.0 & 0.510 & 0.391 & 0.746
  & 53.1 & 0.505 & 0.636 & 0.833 \\
\rowcolor{yellow!20}
GoldModel
  & 90.3 & 0.524 & 0.490 & 0.635
  & 81.6 & 0.505 & 0.569 & $0.587$
  & 83.1 & 0.501 & 0.366 & 0.729
  & 48.3 & 0.511 & 0.369 & $0.765$
  & 53.3 & 0.508 & 0.628 & 0.833 \\
\midrule
Finetuning
  & 88.8 & 0.507 & \textbf{0.493} & 0.714
  & 79.1 & 0.491 & 0.557 & 0.554
  & 82.6 & \textbf{0.497} & 0.381 & 0.733
  & 38.5 & \textbf{0.499} & 0.364 & 0.750
  & 52.3 & 0.505 & 0.619 & 0.829 \\
NegGrad
  & 80.6 & 0.515 & 0.475 & 0.597
  & 76.3 & 0.512 & 0.558 & 0.661
  & 82.5 & \textbf{0.503} & 0.391 & 0.732
  & 45.1 & 0.519 & 0.375 & 0.741
  & 52.7 & 0.504 & 0.628 & 0.829 \\
AdvNegGrad
  & 89.3 & 0.525 & 0.492 & 0.716
  & 69.1 & 0.512 & 0.550 & 0.807
  & 82.6 & 0.495 & 0.376 & 0.731
  & 44.3 & 0.517 & 0.366 & 0.741
  & 51.4 & 0.503 & 0.621 & 0.816 \\
UNSIR
  & 87.2 & 0.512 & 0.489 & 0.867
  & 67.5 & 0.495 & 0.563 & 0.838
  & 82.7 & 0.505 & 0.373 & 0.733
  & 42.8 & 0.521 & 0.382 & 0.749
  & 50.8 & 0.504 & 0.626 & 0.804 \\
BadTeaching
  & \textbf{90.8} & 0.515 & 0.486 & 0.734
  & 82.0 & 0.507 & 0.558 & 0.530
  & 82.9 & \textbf{0.497} & 0.376 & 0.733
  & 48.1 & 0.504 & 0.344 & 0.844
  & 52.6 & 0.504 & 0.628 & 0.815 \\
Scrub
  & 85.4 & 0.497 & 0.484 & 0.605
  & 47.9 & 0.511 & 0.532 & 0.839
  & 82.8 & 0.487 & 0.379 & 0.730
  & 42.4 & 0.522 & 0.388 & 0.726
  & 43.6 & \textbf{0.501} & 0.625 & 0.786 \\
FisherForg.
  & 90.6 & \textbf{0.502} & 0.492 & 0.623
  & 80.4 & \textbf{0.504} & 0.544 & \textbf{0.521}
  & 81.7 & 0.495 & 0.382 & 0.730
  & 48.8 & 0.513 & 0.368 & 0.742
  & \textbf{52.8} & \textbf{0.501} & 0.627 & 0.832 \\
SSD
  & \textbf{90.8} & 0.497 & 0.487 & 0.670
  & 80.3 & 0.510 & 0.549 & 0.535
  & 82.9 & 0.506 & 0.388 & 0.734
  & \textbf{48.9} & 0.510 & 0.381 & 0.743
  & 52.5 & 0.505 & 0.626 & 0.826 \\
SalUn
  & 89.7 & 0.525 & 0.467 & 0.727
  & \textbf{83.8} & 0.514 & 0.541 & 0.543
  & \textbf{83.1} & 0.495 & 0.381 & 0.732
  & 48.2 & 0.506 & 0.356 & 0.855
  & 51.7 & 0.504 & 0.617 & 0.808 \\
\midrule
IDEA
  & \textbf{90.8} & 0.515 & 0.467 & 0.641
  & 79.9 & 0.508 & 0.551 & 0.525
  & 82.4 & 0.599 & \textbf{0.469} & \textbf{0.705}
  & \textbf{48.9} & 0.511 & 0.374 & 0.746
  & 53.3 & 0.504 & 0.626 & 0.833 \\
UtU
  & 91.0 & 0.526 & 0.480 & 0.639
  & 79.7 & 0.500 & 0.564 & 0.525
  & 82.5 & 0.505 & 0.392 & 0.733
  & 49.4 & 0.510 & 0.391 & 0.746
  & 53.3 & 0.505 & 0.636 & 0.833 \\
GNNDelete$^{\S}$
  & 34.6 & 0.520 & 0.450 & \textbf{0.530}
  & 37.9 & 0.516 & \textbf{0.497} & 0.826
  & 75.6 & 0.633 & 0.410 & 0.865
  & 42.2 & 0.496 & \textbf{0.445} & \textbf{0.625}
  & 22.6 & 0.508 & \textbf{0.606} & \textbf{0.682} \\
\midrule
\multicolumn{21}{c}{\textit{SGC architecture}} \\
\midrule
\rowcolor{gray!15}
Original (SGC)
  & 87.3 & 0.483 & 0.419 & 0.538
  & 75.8 & 0.517 & 0.541 & 0.542
  & 71.0 & 0.604 & 0.494 & 0.600
  & 47.8 & 0.511 & 0.082 & 0.902
  & 30.6 & 0.522 & 0.479 & 0.533 \\
\rowcolor{yellow!20}
GoldModel (SGC)
  & 85.7 & 0.485 & 0.427 & 0.542
  & 73.0 & 0.517 & 0.539 & 0.532
  & 70.5 & 0.601 & 0.488 & 0.552
  & 44.9 & 0.502 & 0.059 & 0.906
  & 26.6 & 0.519 & 0.520 & 0.534 \\
\midrule
CGU
  & \textbf{92.4} & \textbf{0.504} & \textbf{0.436} & 0.621
  & \textbf{86.3} & \textbf{0.508} & 0.544 & \textbf{0.512}
  & 75.6 & \textbf{0.594} & \textbf{0.486} & 0.613
  & 5.3 & \textbf{0.495} & \textbf{0.081} & 0.656
  & \textbf{37.4} & 0.536 & 0.545 & \textbf{0.505} \\
ScaleGUN
  & 88.0 & 0.487 & \textbf{0.436} & 0.604
  & 73.7 & 0.509 & \textbf{0.530} & 0.514
  & \textbf{78.7} & 0.596 & 0.478 & \textbf{0.510}
  & \textbf{48.8} & 0.513 & 0.065 & \textbf{0.638}
  & 36.3 & 0.529 & 0.534 & 0.506 \\
CEU
  & 87.3 & \textbf{0.496} & 0.431 & \textbf{0.538}
  & 75.8 & 0.531 & 0.533 & 0.542
  & 71.0 & 0.596 & 0.481 & 0.600
  & 47.8 & 0.507 & 0.059 & 0.902
  & 30.7 & \textbf{0.521} & \textbf{0.512} & 0.533 \\
\bottomrule
\end{tabular}
}
\end{table*}

\subsubsection{New insights from Link Inference Attacks}
\label{sec:rq1}

Table~\ref{tab:rq1} compares the results among Link Inference Attacks and Accuracy.  Across all benchmark datasets, the spread in test accuracy among non-collapsing (i.e., excluding GNNDelete)
unlearning methods is at most 11\, percentage points (pp) on Photos GCN and under 1.5\,pp on DBLP GCN\@.
The vast majority of methods cluster within $\pm$2\,pp of the Original model, making
accuracy unable to distinguish any ranking.
E-UMIA, usually used in traditional MU, is equally blind: across all methods and all datasets it produces AUC $\approx$
0.48--0.53 with no method-discriminating signal, consistent with the finding of
\cite{gu_benchmark} in the node setting.
The sole exception is GNNDelete on DBLP (UMIA $= 0.633$), which reflects how model-collapse
changes the loss distribution rather than actual unlearning. We find that GNNDelete often collapses the model in the EU setting consistently with past findings~\cite{gu_benchmark}. \revision{We define model collapse in terms of accuracy, where the drop is so large that the model is rendered effectively unusable to continue supporting that downstream task.}

\paragraph{LS is a more informative attack.}
Table~\ref{tab:rq1} shows LS AUC across benchmark datasets (GCN,
high centrality setting).
On three of the five datasets, LS produces a spread of at least 20\,pp
across methods, with a consistent ranking structure.
To confirm the qualitative observations above, we test whether the three attacks
differ in their discriminative power across methods.
For each dataset, we compare the variance of each metric across all non-collapsing
methods using one-sided $F$-tests ($H_0$: equal variance; $H_1$: $\sigma^2(\text{LS}) > \sigma^2(\cdot)$).
On AmazonPhotos, AmazonComputers, Flickr, and ogbn-arxiv, the standard deviation of
LS across methods is 6--18$\times$ larger than that of E-UMIA
(7.2$\times$, 17.9$\times$, 6.0$\times$, 7.9$\times$ respectively) and 3--12$\times$ larger
than LT; all eight $F$-tests reject equal variance at $p < 10^{-3}$.
For accuracy, one-sample $t$-tests on $\Delta\text{Acc}$ (deviation from Identity)
yield $p = 0.17$ on AmazonComputers, $p = 0.38$ on DBLP, and $p = 0.07$ on ogbn-arxiv,
confirming that methods produce no significant accuracy change on these datasets.
On AmazonPhotos and Flickr the test is significant ($p < 0.05$), but the mean deviation
is $-2.2$\,pp and $-2.8$\,pp, driven by methods that sacrifice accuracy without any
corresponding reduction in LS, confirming that accuracy variation, where
present, does not reflect unlearning effectiveness.

\rqbox{\textbf{Answer to RQ1:}
Link inference attacks provide substantially more information than accuracy, which is
essentially uninformative across all benchmark datasets.
E-UMIA is uniformly blind.
Link Stealing is the most effective
attack in this setting (high centrality): it produces a consistent, dataset-agnostic ranking with a spread of 20--30\,pp
that accuracy cannot reproduce.
On GCN, only a fraction of the methods reduce LS below the Gold Model, and only at
a non-trivial accuracy cost; most methods either match Identity or are strictly worse.}

\subsubsection{Impact of centrality-based sampling}\label{sec:rq2}

All EU works in the literature sample forget set edges randomly. \our, instead, uses two settings: sampling the lowest and highest centrality edges defined in Section~\ref{sec:difficulty}).  
We report per-method values in Appendix \ref{apx:full_results}.

\begin{wrapfigure}{r}{0.7\linewidth}
    \centering
    \vspace{-5pt}
    \centering    \includegraphics[width=0.8\linewidth]{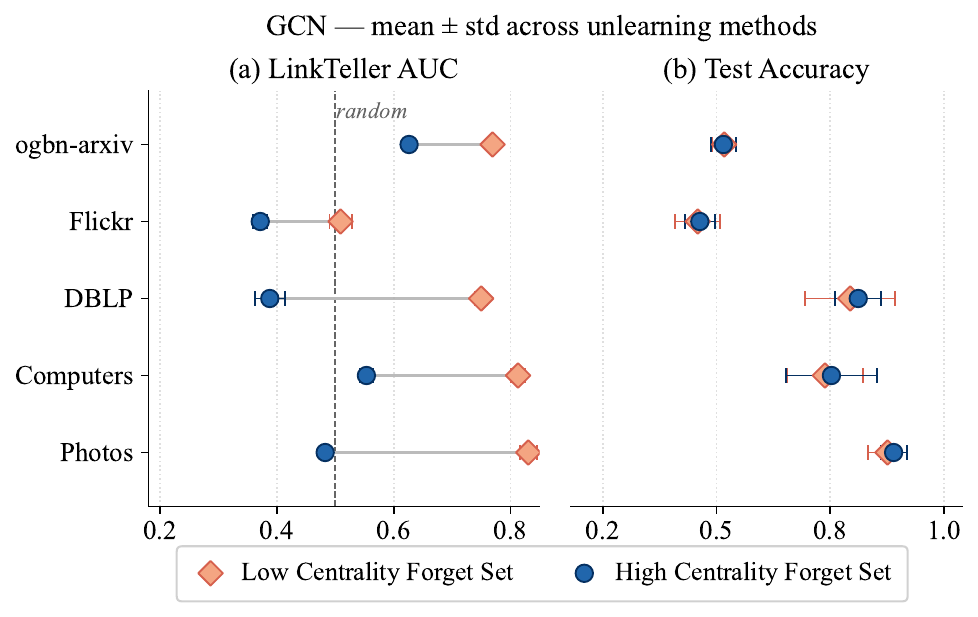}
    \caption{(a) LinkTeller AUC and (b) Test accuracy mean and std across all methods on each dataset when unlearning low centrality (orange) and high centrality (blue) edges.}
    \label{fig:rq2_a}
\vspace{-5pt}
\end{wrapfigure}

On \emph{low-centrality} edges, every method achieves LT AUC $\approx$ \revision{0.75--0.85}, except on Flickr. 
On \emph{high-centrality} edges, LT collapses to a broader range for every method,
landing near the random-chance boundary, but moving between 0.38--0.64 across datasets.
The two settings are shown as connected dot pairs in Figure~\ref{fig:rq2_a}(a): every dumbbell spans roughly 0.35 AUC units.
Figure~\ref{fig:rq2_a}(b) shows test accuracy for the same methods and settings on the same
x-axis scale. For every dataset, the low and high centrality dots on utility nearly overlap. This behavior is explained by how structural information propagates in GNNs.
Low-centrality edges participate in relatively few message-passing paths and thus have a highly \emph{localized} effect on node representations. When such an edge is removed, the change it induces is not compensated by alternative paths, making the absence of the edge clearly detectable by link inference attacks. 
High-centrality edges lie on many message-passing paths and contribute to a large number of node representations. Due to the smoothing effect of GNN aggregation, their influence is diffused across the graph and partially redundant with other paths. When these edges are removed, their signal is absorbed by the surrounding structure.

To confirm that the low-high centrality LT gap is consistent, we apply a Wilcoxon signed-rank test \cite{friedman-wilcoxon} to the paired LT observations obtained by crossing all
non-collapsing unlearning methods.
The test statistic equals $W = 9591$, the theoretical maximum ($W_{\max} = N(N{+}1)/2$),
because \emph{every one} of the low-centrality LT values exceeds the corresponding high-centrality value
($p < 10^{-23}$).
The same test applied to test accuracy yields $W = 3568$ ($p = 0.14$), confirming that the centrality setting has no detectable impact on
predictive utility metrics such as accuracy.

\rqbox{\textbf{Answer to RQ2:} WalkCentrality-based settings, specifically forget sets composed of low/high centrality edges, highlight two main criticalities in current EU evaluation: \textbf{(i).} Low-centrality forget sets remain highly exposed to LinkTeller for
every method on four of five datasets. Because these edges have a localized and non-redundant contribution to node representations, their removal leaves a detectable footprint that is not mitigated by existing unlearning strategies. \textbf{(ii).} Accuracy alone is not able to detect this discrepancy, as it remains largely unchanged in both settings. This confirms that utility-based metrics fail to capture the structural footprint of edges, and can therefore lead to misleading conclusions.}

\subsubsection{Applicability of certified  methods in practice}

\begin{wrapfigure}{r}{0.5\linewidth}
    \centering
    \includegraphics[width=0.9\linewidth]{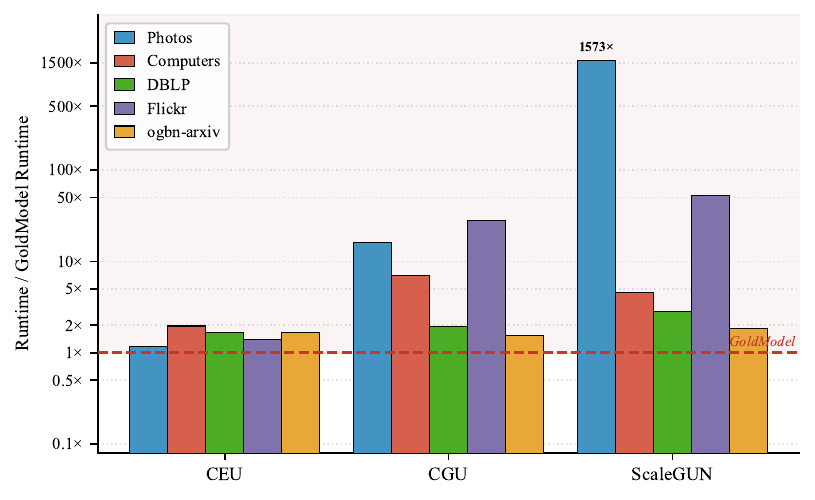}
    \vspace{-5pt}
    \caption{%
      Runtime of certified EU methods relative to the Gold Model (retraining from scratch)
      at $5\%$ forget-set size across all benchmark datasets.
      All bars above the dashed line indicate methods slower than retraining.}
    \label{fig:rq3_a}
\end{wrapfigure}

CEU, CGU, and ScaleGUN all require a linear architecture such as SGC. They provide formal guarantees of closeness to the Gold Model, but as a trade-off, they are not applicable to non-linear GNNs, such as GCNs. 
This is an important limitation, as SGC consistently
underperforms GCN in our benchmark: by $3.6$\,pp on AmazonPhotos, $4.2$\,pp on AmazonComputers, and
$1.2$\,pp on Flickr (refer to Table \ref{tab:rq1}).
The certified guarantee therefore applies to a model that is weaker than any expressive
architecture, mirroring the same restriction reported for certified node unlearning
by \cite{gu_benchmark}. Moreover, these methods are often tested with very small forget sets, in the order of tens of edges. When stress-tested with up to 5\% of edges as the forget set, they are not scalable in terms of runtime, as shown in Figure \ref{fig:rq3_a}. A runtime higher than the one of the Gold Model defeats the purpose of MU itself, as it is always the ideal solution. 

Certified methods do not show consistent reduction on adversary success on attacks. As shown in Table 3, On Photos, both CGU and ScaleGUN worsen LS relative to Identity by $8.3$\,pp (CGU) and $6.6$\,pp respectively; CGU also worsens it on DBLP ($+1.3$\,pp). The improvements they achieve on Flickr ($-24.6$\,pp and $-26.4$\,pp respectively) come at a severe accuracy cost: CGU collapses to $5.3\%$ test accuracy on Flickr, disqualifying it as a viable solution on that dataset.  All three certified methods increase LT relative to Identity on ogbn-arxiv (CEU $+3.3$\,pp, CGU $+6.6$\,pp, ScaleGUN $+5.5$\,pp), meaning the unlearned model is more susceptible to influence-based link inference than the original model that was never modified. This disconnect arises because the certified guarantee bounds closeness to the Gold Model in parameter space, not in the attack-relevant geometry captured by LT and LS.

\rqbox{\textbf{Answer to RQ3:} Certified EU methods impose a double restriction: they require SGC, a linear architecture that doesn't match the performance of non-linear ones, and are only computationally viable when removing a handful of edges. At $5\%$ forget set size, their runtime makes them impractical for all but the smallest datasets, and their formal guarantee does not translate into consistent reduced
adversary success on link inference attacks.}

\section{Conclusion}

In this work, we present \our, a specialized benchmark for Edge Unlearning. \our\ complements existing benchmarks by: \textbf{(i)} Proposing the WalkCentrality metric and using it to implement different forget set selection strategies, \textbf{(ii).} Implementing different link inference attacks to test the reliability of existing GU methods, and \textbf{(iii).} Implementing several general-purpose MU methods and extending them to the graph setting. In particular, we show how the evaluation of EU methods is systematically limited by dataset selection and uninformative metrics for feature-rich datasets. We show that accuracy and UMIA attacks are not informative for the task, and stress the importance of dedicated link inference attacks. We provide a comprehensive benchmark and a modular implementation that can be easily extended with other baselines, datasets, or measures. So far, no previous effort has had such a breadth of inference attacks, but we encourage the community to expand the evaluation. 

\revision{As link inference attacks will get better over time, we need a benchmark that is both up-to-date and easily extensible. With \our, we tested state-of-the-art Link Inference attacks, but we could not possibly cover all existing attacks, or all downstream tasks for GNNs. At the same time, our aim is to give the community a benchmark that will adapt to future work. For this reason, as shown in Appendix \ref{app:running}, \our is easily extendible to other datasets, models, tasks and even alternative negative-set constructions (e.g., held-out edges). Particular care was put into making the code modular and extensible towards this goal.}

\begin{ack}

This work was supported by Danish Data Science Academy, which is funded by the Novo Nordisk Foundation (NNF21SA0069429). 

\end{ack}

\bibliographystyle{plain}
\bibliography{ref}


\appendix

\section*{LLM Usage Disclosure}

\noindent During the preparation of this work, the authors used LLMs to correct typos and grammatical mistakes, and to implement basic methods (the code of the methods was implemented manually). After using this tool/service, the authors reviewed and edited the content as needed and take full responsibility for the content of the published article.

\section*{Broader Impact}

\our\ is designed as a defensive evaluation tool: its purpose is to help researchers identify and correct weaknesses in EU methods before those methods are deployed in privacy-sensitive settings.
The link inference attacks it implements are, however, inherently dual-use.
We release the full benchmark and attack implementations publicly to support reproducible research, as their implementation is already publicly available.
We encourage practitioners to treat a low attack AUC as an empirical signal rather than a privacy certificate: it indicates that a specific adversary, operating under the defined threat model, cannot recover the forgotten edges.
All datasets used in this benchmark are publicly available and contain no personally identifiable information.

\section*{Released artifact and data access}
The released artifact consists of the full benchmark codebase, configuration files, model and unlearning implementations, link-inference attack implementations. We refer the reader to the repository. The benchmark datasets themselves are not redistributed here, as they are all publicly available.
The data was only preprocessed to make it compliant with the framework architecture.

\section{Task and Attacks}\label{apx:definitions}

We focus on the 
setting of edge unlearning for node classification on graphs, and formalize the notions of graph learning, unlearning, and certified unlearning.

\spara{Graph learning.}  Given a parameter space $\Theta$, we refer to a Graph Neural Network (GNN) with parameters $\theta \in \Theta$ as $M_{\theta}$. A graph $G$ is a triple $G=(N,E,X)$ with node set $N$, edge set $E\subseteq N\times N$, and node features $X$; $T = \{(i,y_i) : i \in \mathcal{N}_T\}$ is a training set  defined over $G$, where $\mathcal{N}_T \subseteq N$.

Let $A(G,T)$ be a randomized training algorithm that takes as input a graph $G$ and a training set $T$, and outputs the optimal model parameters $\theta^* \in \Theta$ by minimizing the empirical risk over the labeled nodes in $T$: $
\theta^* = \arg\min\limits_{\theta} \mathcal{L}(\theta,G,T) $. For ease of notation, we will refer to $M_{\theta^*}$ as $\model$.

\spara{Edge Unlearning.} At any time after model training, the model holder can receive an unlearning request that specifies a set of edges to be forgotten, called the \textbf{edge forget set} $\forget \subseteq E$. The retain graph $G'$ is the result of the removal of $\forget$ from $G$, i.e., $G' = (N,E \setminus E_f, X)$.

It follows that the ideal solution would be to retrain $\model$ from scratch on $G'$. We refer to this solution as the gold model $\gold = A(G',T)$, parametrized by $\theta_G$ using the same training algorithm as $\model$.

\begin{definition}[Gold model]
Given a randomized training algorithm $A$ and a retain graph $G'$, the gold model $\gold = A(G', T)$ is the model obtained by running $A$ from scratch on $G'$, without any exposure to edges in the forget set.
\end{definition}

The aim of the model holder at this stage is to avoid full retraining, i.e., $\gold$, and instead use an unlearning method to approximate it as closely as possible:

\begin{definition}[Edge Unlearning (EU)]\label{def:GU}
Given a trained model $\model = A(G,T)$, the retain graph $G'$, and the forget set $\forget$, the goal is to find an updated model $M_{\theta_u}$ that closely approximates $\gold$ and leaks no information on the edges in $\forget$.
\end{definition}

In the following, we refer to the unlearned model $M_{\theta_u}$ as simply $\unlearned$.
Clearly, the gold model $\gold$ is the reference output model for almost all metrics in evaluating unlearning.

\textbf{Certified Unlearning.} Certified methods formalize the notion of \textit{closeness} between $\unlearned$ and 
$\gold$ by drawing on the framework of differential privacy \cite{differential_privacy}.
Specifically, an EU method is said to be $(\epsilon, \delta)$-certified if, for every 
forget set $\forget \subseteq E$ and every subset $\Theta' \subseteq \Theta$, we have:
\[  (i) \quad P[\theta_u \in \Theta']  \leq e^\epsilon 
P[\theta_G \in \Theta'] + \delta \]
\[  (ii) \quad P[\theta_G \in \Theta']  \leq e^\epsilon 
P[\theta_u \in \Theta'] + \delta \]
Together, conditions (i) and (ii) impose a bidirectional bound on the distributional 
discrepancy between $\theta_u$ and $\theta_G$. 
Intuitively, $\epsilon$ controls how tightly the two distributions must agree, while 
$\delta$ allows for a small relaxation. This definition is adapted 
from~\cite{IDEA,cgu,ceu}.

\subsection{Link Inference Attacks}
\label{apx:attacks}

While accuracy-based metrics ensure that $M_u$ preserves predictive performance, they are largely insensitive to edge removals if node features are informative (Section~\ref{sec:flaws_current_evaluations}).
To directly assess whether the forgotten edge information has been removed from the model, \our\ adopts three established \emph{link inference attacks}, whereby an adversary with black-box access to $M_u$ queries the model to determine whether a specific edge $e \in E_f$ was present during training. The goal is to standardize evaluation of the resilience of unlearning methods to adversaries attempting to recover forgotten information.

\textbf{Threat model.}
We consider a membership inference adversary that has access to the unlearned model $M_u$ and to the retain graph $G' = (N, E \setminus E_f, X)$, but not to the original graph $G$ or to $E_f$.
The adversary must classify a candidate edge $(u, v)$ as either a forgotten edge (positive) or a non-existing edge (negative).
Crucially, all attacks query $M_u$ using $G'$ as the message-passing graph: providing $G$ would trivially inflate the scores for edges in $E_f$ through direct propagation, regardless of what the model has actually learned.
A well-unlearned model should be indistinguishable from the gold model $M_G = A(G', T)$, making forgotten edges look like non-existing ones; we report AUC (or accuracy) relative to the 0.5 random-chance baseline.

\textbf{Positive and negative sets.}
For each attack, the positive set consists of the forgotten edges $E_f$, deduplicated to undirected pairs.
The negative set is drawn uniformly at random from all non-existing pairs $(u, v) \notin E$, balanced to match $|E_f|$.
This unbiased sampling avoids introducing structural confounders, such as selecting non-edges ($u$,$v$) where $u$ and $v$ have many shared neighbours that would systematically inflate their scores relative to forgotten edges whose direct connection has been removed.

\paragraph{Edge Membership Inference Attack (\textsc{UMIA}).}
We adapt the population membership inference attack of~\cite{scrub} to the edge setting.
Because edge removal affects nodes indirectly through the message-passing neighbourhood rather than directly, we operate at the node level on the set of \emph{infected nodes} (see Section \ref{sec:flaws_current_evaluations}).
These are the nodes whose $k$-hop neighbourhood overlaps with at least one endpoint of a forgotten edge, and whose learned representations may carry residual information about $E_f$.

The attack proceeds as follows.
\begin{enumerate*}[label=(\arabic*)]
\item Run $M_u$ on $G'$ and collect the per-node cross-entropy loss for every node.
\item Assign label $1$ to infected nodes in the forget set $\mathcal{N}_f$ and label $0$ to test nodes $\mathcal{N}_T$.
\item Train a logistic regression classifier on the scalar losses using 3-fold stratified cross-validation
\item Report the mean accuracy.
\end{enumerate*}
An accuracy of $0.5$ indicates perfect unlearning: the loss distribution over infected nodes is indistinguishable from the test distribution.

\paragraph{LinkTeller (LT).}
LinkTeller \cite{linkteller} infers edge existence by measuring the \emph{influence} of node $v$'s features on node $u$'s output embedding.
If the model retains memory of a forgotten edge $(u, v)$, perturbing $v$'s input features will produce a larger change in $u$'s output than for a truly non-existing pair, because the learned parameters still encode the dependency introduced during training on $G$.

Formally, the influence score is the directional derivative of $u$'s embedding with respect to $v$'s features, in the direction of $v$'s own feature values, approximated via a first-order finite difference:
\begin{equation}
    s(u, v) = \left\lVert \frac{M_u(X + \varepsilon \cdot \mathbf{x}_v \mathbf{e}_v^\top,\, G')[u] \;-\; M_u(X, G')[u]}{\varepsilon} \right\rVert_2,
\end{equation}
where $\mathbf{e}_v$ is the standard basis vector for node $v$, so that $\mathbf{x}_v \mathbf{e}_v^\top$ perturbs only row $v$ of the feature matrix by a fraction $\varepsilon$ of its own value.
Both model calls use $G'$ as the edge index.
We compute $s(u,v)$ for all positive pairs $(u,v) \in E_f$ and all negative pairs, and report the ROC-AUC of the resulting scores.
An AUC of $0.5$ indicates that the model assigns equivalent influence scores to forgotten and non-existing edges.

\paragraph{Link Stealing Attack (LS).}
Link stealing attacks \cite{stealing_links} exploit the observation that a GNN trained on a graph tends to produce more similar output distributions for pairs of nodes that are connected than those that are not.
If unlearning is effective, the posteriors of forgotten edge endpoints should become as dissimilar as those of non-existing pairs.

Given the unlearned model $M_u$ evaluated on $G'$, we compute the softmax class posteriors $p_v = \mathrm{softmax}(M_u(X, G')[v]) \in \mathbb{R}^C$ for every node $v$, where $C$ is the number of classes.
For each candidate pair $(u, v)$, the attack score is the cosine distance between posteriors:
\begin{equation}
    s(u, v) = 1 - \frac{p_u \cdot p_v}{\lVert p_u \rVert_2\, \lVert p_v \rVert_2}.
\end{equation}
We report the ROC-AUC of these scores over positive (forgotten) and negative (non-existing) pairs, symmetrised to be $\geq 0.5$.
An AUC of $0.5$ indicates that the model's posteriors no longer reveal membership of edges in $E_f$.

\subsection{Infected nodes study}\label{apx:infected}

As shown in Table~\ref{tab:infected_nodes}, removing even 1\% of edges on these datasets infects 58--76\% of all nodes within two hops, meaning that MU methods are effectively operating on almost the entire graph, leaving little room to amortise the cost of unlearning against the cost of retraining.  When the forget set grows to 5\% of edges, the infected set of nodes within two hops represent  more than 85\% of nodes.

\begin{table}[t]
\centering
\caption{Fraction of nodes (\%) affected by 2-hop neighbourhood expansion
         when removing a random subset of edges. Values averaged over
         5 random forget-set samples.
         \textbf{Bold} entries exceed 80\%.}
\label{tab:infected_nodes}
\scriptsize
\setlength{\tabcolsep}{4pt}
\begin{tabular}{@{}l rr rrrrr@{}}
\toprule
Dataset & $|V|$ & $|E|$ & 1\% & 5\% & 10\% & 20\% & 50\% \\
\midrule
Cora & 2,708 & 5,278 & 58.0{\scriptsize$\pm$2.3} & \textbf{85.0{\scriptsize$\pm$0.5}} & \textbf{90.7{\scriptsize$\pm$0.6}} & \textbf{94.2{\scriptsize$\pm$0.3}} & \textbf{97.6{\scriptsize$\pm$0.4}} \\
Citeseer & 3,327 & 4,552 & 26.3{\scriptsize$\pm$1.3} & 53.1{\scriptsize$\pm$0.9} & 65.1{\scriptsize$\pm$0.4} & 75.9{\scriptsize$\pm$0.3} & \textbf{88.5{\scriptsize$\pm$0.2}} \\
Pubmed & 19,717 & 44,324 & 76.2{\scriptsize$\pm$1.4} & \textbf{94.5{\scriptsize$\pm$0.3}} & \textbf{97.8{\scriptsize$\pm$0.2}} & \textbf{99.3{\scriptsize$\pm$0.1}} & \textbf{$\approx$100{\scriptsize$\pm$0.0}} \\
\bottomrule
\end{tabular}
\end{table}

\section{WalkCentrality experiments}\label{apx:walkcentrality}

In this section we show two empirical experiments to support the definition of WalkCentrality.

\revision{\paragraph{(i) Redundancy absorbs removal.}
We compute the correlation between WalkCentrality and the increase in training loss when the corresponding edges are removed, that is, how much the model's fit depended on them. The correlation scores are shown in Table \ref{tab:apx_corr}.}
˙
\begin{table}[t]
    \centering
    \caption{Spearman correlation between WalkCentrality and the increase in training loss after edge removal.}
    \label{tab:walkcentrality_loss}
    \begin{tabular}{lc}
        \toprule
        Dataset & $\rho$ (WalkCentrality, loss increase) \\
        \midrule
        Photos    & \textbf{-0.989} \\
        Computers & \textbf{-0.993} \\
        DBLP      & \textbf{-0.989} \\
        \bottomrule
    \end{tabular}
    \label{tab:apx_corr}
\end{table}

\revision{The strong negative correlation confirms that the higher an edge's WalkCentrality, the more completely the rest of the graph absorbs its removal.}

\revision{\paragraph{(ii) Low-centrality removals are localized and large.}
For a removed edge $e=(u,v)$, let
\[
d_i = \left\lVert h_i(G) - h_i(G \setminus e) \right\rVert_2
\]
be how far node $i$'s hidden representation moves (64-dimensional post-ReLU layer). We report
\[
m(e) = \sqrt{d_u^2 + d_v^2},
\]
the displacement of the removed edge's own two endpoints, and give the median of $m(e)$ over the edges of each forget set. Results are shown in Table \ref{tab:apx_displacement}.}

\begin{table}[t]
    \centering
    \caption{Endpoint displacement after edge removal for low- and high-WalkCentrality edges.}
    \label{tab:walkcentrality_displacement}
    \begin{tabular}{lccc}
        \toprule
        Dataset & Low WalkCentrality & High WalkCentrality & Ratio \\
        \midrule
        Photos    & 0.309 & 0.060 & \textbf{5.2$\times$} \\
        Computers & 0.274 & 0.042 & \textbf{6.7$\times$} \\
        DBLP      & 0.289 & 0.056 & \textbf{5.3$\times$} \\
        \bottomrule
    \end{tabular}
    \label{tab:apx_displacement}
\end{table}

\revision{The two results form a single chain: high-WalkCentrality edges are redundant (i), so their removal displaces their endpoints only slightly (ii). LinkTeller measures exactly this endpoint influence, which is why it reaches 0.73--0.85 in the low-WalkCentrality setting and falls to 0.38--0.64 in the high-WalkCentrality one (Figure \ref{fig:rq2_a}(a)).
}

\revision{\section{Ablation Studies}\label{apx:ablations}}


\subsection{Ablation on the attacker's accessible information}\label{apx:attacker_ablation}

In this ablation study, we ablate the adversary's accessible information across four levels for LinkStealing (LS): we ablate the attacker's access to the unlearned model + 0\%, 25\%, 50\%, 100\% of the retained graph (the 100\% setting is the one in the paper) on GCN.

We tested this ablation on both Photos and Flickr, the datasets with the highest LS spread in the paper, across all unlearning methods. 

\begin{table}[t]
\centering
\caption{LinkStealing ablation on attacker's access.}
\label{tab:attacker_ablation}
\begin{tabular}{lccc}
\toprule
Method & None & 25\% & 50\% \\
\midrule
Identity     & 0.518 & 0.641 & 0.639 \\
Gold Model   & 0.547 & 0.673 & 0.672 \\
Finetuning   & 0.522 & 0.636 & 0.634 \\
NegGrad      & 0.564 & 0.681 & 0.678 \\
Adv.NegGrad  & 0.554 & 0.685 & 0.682 \\
UNSIR        & 0.776 & 0.861 & 0.863 \\
BadTeaching  & 0.581 & 0.699 & 0.696 \\
Scrub        & 0.662 & 0.757 & 0.758 \\
Fisher       & 0.521 & 0.653 & 0.651 \\
SSD          & 0.513 & 0.626 & 0.624 \\
SalUn        & 0.609 & 0.730 & 0.726 \\
\textbf{spread} & \textbf{26 pp} & \textbf{23 pp} & \textbf{24 pp} \\
\bottomrule
\end{tabular}
\end{table}

Table \ref{tab:attacker_ablation} shows the results for this ablation. Notice how, uniformly for each method, the attacker's success increases as they gain access to more information. However, different percentages of the retain graph behave closely.

Table \ref{tab:f-tests_ablation} shows that this spread is consistent on both datasets, with one-sided
$F$-test $\sigma^2(\mathrm{LS})>\sigma^2(\text{E-UMIA})$ in brackets.

\begin{table}[t]
\centering
\caption{F-tests on LinkStealing attacker's access ablation}
\label{tab:f-tests_ablation}
\begin{tabular}{lcc}
\toprule
Access to $G'$ & Photos & Flickr \\
\midrule
0\%   & 26 pp ($p{=}3\times10^{-9}$)  & 2 pp ($p{=}0.4$) \\
25\%  & 23 pp ($p{=}2\times10^{-8}$)  & 6 pp ($p{=}3\times10^{-3}$) \\
50\%  & 24 pp ($p{=}2\times10^{-8}$)  & 11 pp ($p{=}1\times10^{-5}$) \\
100\% & 24 pp ($p{=}1\times10^{-8}$)  & 14 pp ($p{=}2\times10^{-6}$) \\
\bottomrule
\end{tabular}
\end{table}

At every level with any structural access it separates methods significantly, where accuracy (spread within ±7 pp of the original model on flickr and ±12 on Photos) and E-UMIA do not. So, across all settings, LS is significantly more informative than both E-UMIA and Accuracy. 

\rqbox{Because of this, we can conclude that LS discriminates better than accuracy and E-UMIA even when the attacker has less information, showing that the answer to RQ1 is robust.} 

The conclusion is therefore stable to this ablation, although attack effectiveness decreases substantially when structural information is unavailable (as expected). 

\subsection{Other forget set selection strategies}\label{apx:forget_strategy}

To show that WalkCentrality forget sets act as difficulty bounds, we introduce two new, more realistic selection strategies, and compare them.

\begin{itemize}
\item \textbf{Community-based forgetting:} We simply pick a class of users whose edges we will be removing. For Photos, that is a product category. For DBLP, it is a research community. For arxiv, it is a subject area. Both endpoints need to be in the same class.
\item \textbf{Time-based forgetting:} We use arxiv and date each edge as the max among the publication year of the endpoints (a citation can't exist until the later paper is published). We order the edges, and we remove all the older ones to reach 5\%.
\end{itemize}

All experiments are run with a GCN on a 5\% forget set and across all unlearning methods.

For scenario 1, we show that the accuracies and E-UMIA are still uninformative with respect to our settings. This is reported in Table \ref{tab:centrality_community}.

\begin{table}[t]
\centering
\caption{Metrics on the community-built forget set on DBLP.}
\label{tab:centrality_community}
\begin{tabular}{lccc}
\toprule
Metric & Low Centrality & High Centrality & Community \\
\midrule
Accuracy (test)   & 0.8802 & 0.8873 & 0.8699 \\
Accuracy (forget) & 0.8901 & 0.8964 & 0.8548 \\
E-UMIA            & 0.5162 & 0.5128 & 0.4981 \\
\bottomrule
\end{tabular}
\end{table}

Table \ref{tab:method_centrality_community} shows the Link Teller value on a subset of unlearning methods. It is easy to see that, for all methods (except GNNdelete, who is also collapsing the model), the community Link Teller is between the low and high centrality settings. Basically, our settings are stress-testing the methods; realistic forget sets, like from a community, are likely to land between our values, because they contain some low centrality edges along with high centrality ones.

\begin{table}[t]
\centering
\caption{Link Teller values for the community-built forget set on DBLP.}
\label{tab:method_centrality_community}
\begin{tabular}{lccc}
\toprule
Method & Low Centrality & High Centrality & Community \\
\midrule
Original     & 0.7480 & 0.4535 & 0.6786 \\
GoldModel    & 0.7486 & 0.4589 & 0.6773 \\
Finetuning   & 0.7481 & 0.4717 & 0.6756 \\
NegGrad      & 0.7788 & 0.4487 & 0.6701 \\
AdvNegGrad   & 0.7491 & 0.4600 & 0.6697 \\
UNSIR        & 0.7572 & 0.4626 & 0.6715 \\
BadTeaching  & 0.7492 & 0.4571 & 0.6769 \\
Scrub        & 0.7473 & 0.4573 & 0.6768 \\
FisherForg.  & 0.7424 & 0.4545 & 0.6764 \\
SSD          & 0.7436 & 0.4658 & 0.6788 \\
SalUn        & 0.7545 & 0.4481 & 0.6659 \\
GNNDelete    & 0.6630 & 0.4329 & 0.3982 \\
\bottomrule
\end{tabular}
\end{table}

For scenario 2, on arxiv, the same claims hold. Accuracy and E-UMIA are uninformative, as shown in Table \ref{tab:time_forget_metrics}. 

\begin{table}[t]
\centering
\caption{Performance across different node groups and temporal windows.}
\label{tab:time_forget_metrics}
\begin{tabular}{lccc}
\toprule
Metric & Low Centrality & High Centrality & Temporal Window \\
\midrule
Accuracy (test)   & 0.5168 & 0.5151 & 0.5138 \\
Accuracy (forget) & 0.5276 & 0.5262 & 0.5220 \\
E-UMIA            & 0.5051 & 0.5040 & 0.5020 \\
\bottomrule
\end{tabular}
\end{table}

The realistic, time-based selection of forget set lands in the middle of the two centrality settings in terms of Link Teller, as shown in \ref{tab:time_forget_comparison}, although the gap is much narrower with respect to other datasets.

\begin{table}[t]
\centering
\caption{Performance across different node groups and temporal windows for each method.}
\label{tab:time_forget_comparison}
\begin{tabular}{lccc}
\toprule
Method & Low Centrality & High Centrality & Temporal Window \\
\midrule
Original     & 0.7906 & 0.6357 & 0.7179 \\
GoldModel    & 0.7707 & 0.6284 & 0.7185 \\
Finetuning   & 0.7677 & 0.6191 & 0.7183 \\
NegGrad      & 0.7665 & 0.6275 & 0.7150 \\
AdvNegGrad   & 0.7622 & 0.6211 & 0.7205 \\
UNSIR        & 0.7681 & 0.6264 & 0.7154 \\
BadTeaching  & 0.7677 & 0.6279 & 0.7141 \\
Scrub        & 0.7659 & 0.6246 & 0.7192 \\
FisherForg.  & 0.7681 & 0.6275 & 0.7134 \\
SSD          & 0.7664 & 0.6256 & 0.7185 \\
\bottomrule
\end{tabular}
\end{table}

\rqbox{Accuracy and UMIA are equally uninformative when the forget set selection is not random, but selected through different means, like Community or Time. In these cases, the Link Teller values fall between the low and high WalkCentrality settings, which correctly act as "difficulty" bounds for link inference retrieval.}

\subsection{Results on Link Prediction}\label{apx:link_prediction}

All of the graph-based Unlearning methods that we include in our benchmark were designed and tested for node classification. Hence, \our is also benchmarked on node classification. In this Section, we ablate the downstream task of link prediction to \our and run it on Photos and DBLP with GCN, under both WalkCentrality forget-set settings (high and low centrality).

To be as faithful as possible to the main text experiments, the encoder is the same GCN; a decoder scores a node pair $(u,v)$ from the two endpoint embeddings, $s(u,v)=\mathrm{MLP}([\mathbf{z}_u \Vert \mathbf{z}_v])$. 

We would like to clarify that, even in this setting, WalkCentrality is unchanged. It is defined on $\hat{A}$, a property of the graph, independent of the task or the model. LinkTeller also remains unchanged, and Link Stealing needs only one change: the pairwise distance is taken on the node embeddings instead of on softmax-normalised posteriors. The attack logic is otherwise identical.

\begin{table}[t]
\centering
\caption{Link prediction performance on Photos and DBLP across low- and high-centrality forget sets. LP = Link Prediction. LT = Link Teller. Low = Low WalkCentrality set. High = High WalkCentrality set.}
\label{tab:lp_centrality}
\begin{tabular}{lcccccccc}
\toprule
& \multicolumn{4}{c}{Photos} & \multicolumn{4}{c}{DBLP} \\
\cmidrule(lr){2-5} \cmidrule(lr){6-9}
Method & \multicolumn{2}{c}{Low} & \multicolumn{2}{c}{High}
       & \multicolumn{2}{c}{Low} & \multicolumn{2}{c}{High} \\
\cmidrule(lr){2-3} \cmidrule(lr){4-5}
\cmidrule(lr){6-7} \cmidrule(lr){8-9}
& LP AUC & LT & LP AUC & LT
& LP AUC & LT & LP AUC & LT \\
\midrule
Original     & 0.960 & 0.947 & 0.956 & 0.929
             & 0.856 & 0.757 & 0.849 & 0.719 \\
Gold Model   & 0.955 & 0.949 & 0.813 & 0.932
             & 0.859 & 0.741 & 0.851 & 0.674 \\
Finetuning   & 0.944 & 0.959 & 0.929 & 0.922
             & 0.855 & 0.736 & 0.847 & 0.724 \\
NegGrad      & 0.947 & 0.957 & 0.937 & 0.922
             & 0.855 & 0.736 & 0.848 & 0.724 \\
Adv. NegGrad & 0.945 & 0.958 & 0.941 & 0.924
             & 0.855 & 0.736 & 0.848 & 0.724 \\
GNNDelete    & 0.747 & 0.940 & 0.748 & 0.900
             & 0.731 & 0.761 & 0.589 & 0.698 \\
\bottomrule
\end{tabular}
\end{table}

\begin{table}[t]
\centering
\caption{Mean LS and LT values for Link prediction.}
\label{tab:mean_lp_tab}
\begin{tabular}{llcc}
\toprule
Attack & Forget Set & Photos & DBLP \\
\midrule
Link Stealing & Low Centrality  & 0.950 & 0.860 \\
Link Stealing & High Centrality & 0.888 & 0.812 \\
LinkTeller    & Low Centrality  & 0.954 & 0.741 \\
LinkTeller    & High Centrality & 0.926 & 0.713 \\
\bottomrule
\end{tabular}
\end{table}

The resulting link prediction models are strong, so the evaluation is meaningful: the original model reaches an AUC of $0.960$ on Photos, $0.856$ on DBLP on the test set.

We then measure both link inference attacks under low- and high-centrality forget sets, exactly as in Section \ref{sec:experiments}.  We also report the same statistical tests.  Results are shown in Table \ref{tab:lp_centrality}, and the mean across all methods is reported in Table \ref{tab:mean_lp_tab}.

Applying the same one-sided Wilcoxon signed-rank test used in the paper, paired over methods, gives  $p = 3\times10^{-5}$ for Link Stealing and $p = 1\times10^{-3}$ for LinkTeller when the comparisons are pooled. This is exactly coherent with the Node classification results we show in the paper. However, as shown in the table, the magnitude of the gap is smaller relative to node classification.

\rqbox{Both WalkCentrality and the \our pipeline effortlessly generalize to link prediction and similar downstream tasks. The RQ2 conclusion replicates on link prediction: low-centrality forget sets are more exposed to link inference than high-centrality ones, for every method, on every dataset, under both Link Stealing and Link Teller: all paired comparisons agree in direction.}

Finally, the collapse reported in the paper reproduce on the new task: GNNDelete collapses under link prediction as well, losing between 12 and 52 AUC points relative to the original model depending on dataset and setting, consistent with our node classification experiments and with \cite{gu_benchmark}.

\revision{\section{Methods}\label{apx:methods}}

\begin{table*}[t]
\centering
\caption{Overview of the unlearning methods considered in this work.
``Graph-native'' denotes methods that were originally proposed for and
evaluated on graph neural networks (GNNs).}
\label{tab:method_overview}
\resizebox{\textwidth}{!}{%
\begin{tabular}{l l c c c c}
\toprule
Method & Family & Graph-native & Certified & Needs retain set & Needs forget set \\
\midrule
Original ($M^*$) &
Reference &
--- & --- & \xmark & \xmark \\

Retrain / Gold ($M_G$) &
Reference &
--- & Exact & \cmark & \xmark \\

Fine-Tuning &
Training &
\xmark & \xmark & \cmark & \xmark \\

NegGrad &
Gradient ascent &
\xmark & \xmark & \xmark & \cmark \\

AdvancedNegGrad (NegGrad+) &
Gradient ascent + descent &
\xmark & \xmark & \cmark & \cmark \\

UNSIR &
Noise / impair--repair &
\xmark & \xmark & \cmark & \cmark \\

Bad Teaching &
Knowledge distillation &
\xmark & \xmark & \cmark & \cmark \\

SCRUB &
Knowledge distillation &
\xmark & \xmark & \cmark & \cmark \\

Fisher Forgetting &
Weight scrubbing (Fisher) &
\xmark & \xmark & \cmark & \xmark \\

SSD &
Weight dampening (Fisher) &
\xmark & \xmark & \cmark{} (train) & \cmark \\

SalUn &
Saliency mask + random labels &
\xmark & \xmark & \cmark & \cmark \\

GNNDelete &
Structural / deletion operator &
\cmark & \xmark & \xmark & \cmark \\

CGU &
Influence function &
\cmark & $\cmark$ ($\varepsilon,\delta$) & \cmark & \cmark \\

CEU &
Influence function &
\cmark & $\cmark$ ($\varepsilon,\delta$) & \cmark & \cmark \\

IDEA &
Influence function &
\cmark & \cmark{} (instance-level) & \cmark & \cmark \\

ScaleGUN &
Influence function + approx.\ propagation &
\cmark & $\cmark$ ($\varepsilon,\delta$) & \cmark & \cmark \\

UnlinkToUnlearn &
Propagation on retain graph &
\cmark & \xmark & \xmark & \cmark \\
\bottomrule
\end{tabular}%
}
\end{table*}

\paragraph{Original and Gold Model.}
The Original model ($M^*$) is the model before unlearning and serves as a
reference point for evaluating utility and forgetting. The Gold model ($M_G$)
is obtained by retraining from scratch after removing the forget set and
therefore represents the exact retraining baseline.

\paragraph{Fine-Tuning.}
Fine-Tuning adapts the pretrained model using the available training data
without explicitly optimizing against the forget set. It provides a simple
baseline for evaluating whether standard model adaptation can achieve
unlearning.

\paragraph{NegGrad and AdvancedNegGrad.}
NegGrad performs gradient ascent on the forget set to increase the loss on
forgotten examples. AdvancedNegGrad (NegGrad+) combines this forgetting
objective with gradient descent on retained data, aiming to preserve model
utility while degrading performance on the forget set.

\paragraph{UNSIR.}
UNSIR uses an impair--repair procedure in which the model is first impaired
using noise-based perturbations and subsequently repaired using retained
data. The procedure aims to remove information associated with the forget
set while recovering performance on retained examples.

\paragraph{Bad Teaching and SCRUB.}
Bad Teaching uses knowledge distillation to train a student model while
suppressing information associated with the data to be forgotten. SCRUB
similarly uses a teacher--student distillation framework, combining
forgetting and retention objectives to remove information while maintaining
utility.

\paragraph{Fisher Forgetting and SSD.}
Fisher Forgetting uses the Fisher information matrix to identify parameters
that are important for the forget set and performs weight scrubbing.
SSD (Selective Synaptic Dampening) similarly uses Fisher-information-based
parameter importance, selectively dampening parameters associated with the
data to be forgotten.

\paragraph{SalUn.}
SalUn uses parameter saliency to identify model weights associated with the
forget set and applies a saliency-guided unlearning procedure involving
random-label training.

\paragraph{GNNDelete.}
GNNDelete is a graph-native unlearning method that inserts a trainable,
layer-wise deletion operator into the GNN, which transforms the node
representations in the vicinity of the deleted nodes or edges while the
original model weights remain frozen. It is optimized to jointly satisfy
two properties: consistency on deleted edges and preserved influence on
neighboring nodes.

\paragraph{CGU and CEU.}
CGU and CEU are graph-native methods based on influence-function
approximations. They estimate the effect of removing training instances on
the model and use this information to update the GNN without full
retraining. Both provide $(\varepsilon,\delta)$-style certified guarantees
under their respective assumptions.

\paragraph{IDEA.}
IDEA is a graph-native influence-function-based unlearning method that
approximates the effect of individual training instances on the model.
It provides an instance-level certified guarantee while avoiding complete
retraining.

\paragraph{ScaleGUN.}
ScaleGUN combines influence-function-based parameter updates with
approximate propagation mechanisms designed for scalability on graphs. It
provides $(\varepsilon,\delta)$-style certified guarantees while reducing
the computational cost of graph unlearning.

\paragraph{UnlinkToUnlearn.}
UnlinkToUnlearn performs graph-native unlearning by removing forget-set
edges from the input graph and simply re-running inference on this retain
graph with the original, unmodified model weights. It requires no
parameter update and no explicit retain set, relying instead on the
absence of the forgotten edges to block their influence during message
passing.

\section{Running Experiments with \our}
\label{app:running}

\our\ follows a fully configuration-driven design similarly to ERASURE \cite{erasure}, but completely rebuilding the framework from scratch to support graphs.
All experiments are launched through a single entry point:

\begin{lstlisting}[language=bash]
python main.py <config>.jsonc
\end{lstlisting}

Each configuration file (JSONC format) fully specifies an experiment through four top-level sections.

\paragraph{Data.}
The \texttt{data} block defines the dataset and partitioning strategy.
The graph is split into \texttt{train}, \texttt{validation}, \texttt{test}, \texttt{forget}, and \texttt{retain} partitions.
The forget set is constructed by a centrality-aware splitter whose \texttt{mode} field selects either low-centrality or high-centrality edges, and whose \texttt{percentage} controls the forget-set size (e.g.\ $5\%$).
A fixed \texttt{split\_seed} ensures reproducibility across runs.

\paragraph{Predictor.}
The \texttt{predictor} block specifies the GNN architecture (e.g., \textsc{GCN}, \textsc{GIN}, \textsc{GAT}, \textsc{GraphSAGE}, \textsc{SGC}) together with training hyperparameters such as learning rate, number of epochs, and loss function.
A unique \texttt{alias} identifies the model for caching: the original model $M^*$ and the Gold Model $M_G$ are trained once and reused across all unlearning methods in the same configuration.

\paragraph{Unlearners.}
The \texttt{unlearners} field is a list of unlearning methods to evaluate sequentially.
Each entry specifies a method class and its hyperparameters.
The Identity baseline (original model, no unlearning) and the Gold Model (full retraining on the retain set) are always included as reference points.
Adding a new method requires only appending a new entry to this list.

\paragraph{Evaluator.}
The \texttt{evaluator} block enumerates the metrics to compute for each method: runtime, node classification accuracy (test, forget, and train splits), macro-F1 score, E-UMIA, LinkTeller, and Link Stealing.
Results are serialised to a JSON file for downstream analysis and aggregation.

\paragraph{Configurations.}
All configurations used in our benchmark are provided under \texttt{configs/benchmark/lethe/}, with one file per (dataset, architecture, centrality-setting) triple.
Extending \our\ to a new dataset or method requires writing a new configuration file or appending to an existing one; no changes to the framework code are needed.

\subsection{Extending \our}
\label{app:extending}

\our\ is designed to be modular: every component is resolved at runtime from the \texttt{class} and \texttt{parameters} fields of the configuration file, so adding a new dataset, architecture, unlearning method, or attack requires no changes to the framework core.

\paragraph{New dataset.}
Any dataset available in PyTorch Geometric can be included by pointing the \texttt{data} block at the corresponding class:
\begin{lstlisting}[language=json]
"datasource": {"class": "torch_geometric.datasets.MyDataset",
               "parameters": {"root": "resources/data"}}
\end{lstlisting}
Custom datasets can be integrated by wrapping them in a \texttt{TorchGeometricDataSource}-compatible class that exposes the standard PyG \texttt{Data} interface.

\paragraph{New GNN architecture.}
A new architecture is a standard \texttt{torch.nn.Module} placed under \texttt{lethe/model/graphs/}.
It only needs to implement a \texttt{forward(x, edge\_index)} signature.
Once defined, it is referenced in the \texttt{predictor} block of any configuration file without further registration.

\paragraph{New unlearning method.}
A new unlearning method subclasses \texttt{GraphUnlearner} and implements the unlearning logic.
The base class provides access to the trained model, the forget and retain partitions, and the infected-node computation.
The new class is then added to the \texttt{unlearners} list of any configuration file:
\begin{lstlisting}[language=json]
{"class": "path.to.MyUnlearner", "parameters": {...}}
\end{lstlisting}

\paragraph{New link inference attack.}
A new attack subclasses \texttt{GraphMeasure} and implements a \texttt{process(evaluation)} method that computes a scalar score per candidate edge pair and records results via \texttt{evaluation.add\_value}.
It is added to the \texttt{evaluator.measures} list exactly like the built-in attacks (UMIA, LinkTeller, Link Stealing), with no additional wiring required.

\section{Complementary Results}\label{apx:full_results}

\begin{figure}[p]
    \centering
    \includegraphics[width=0.9\textwidth]{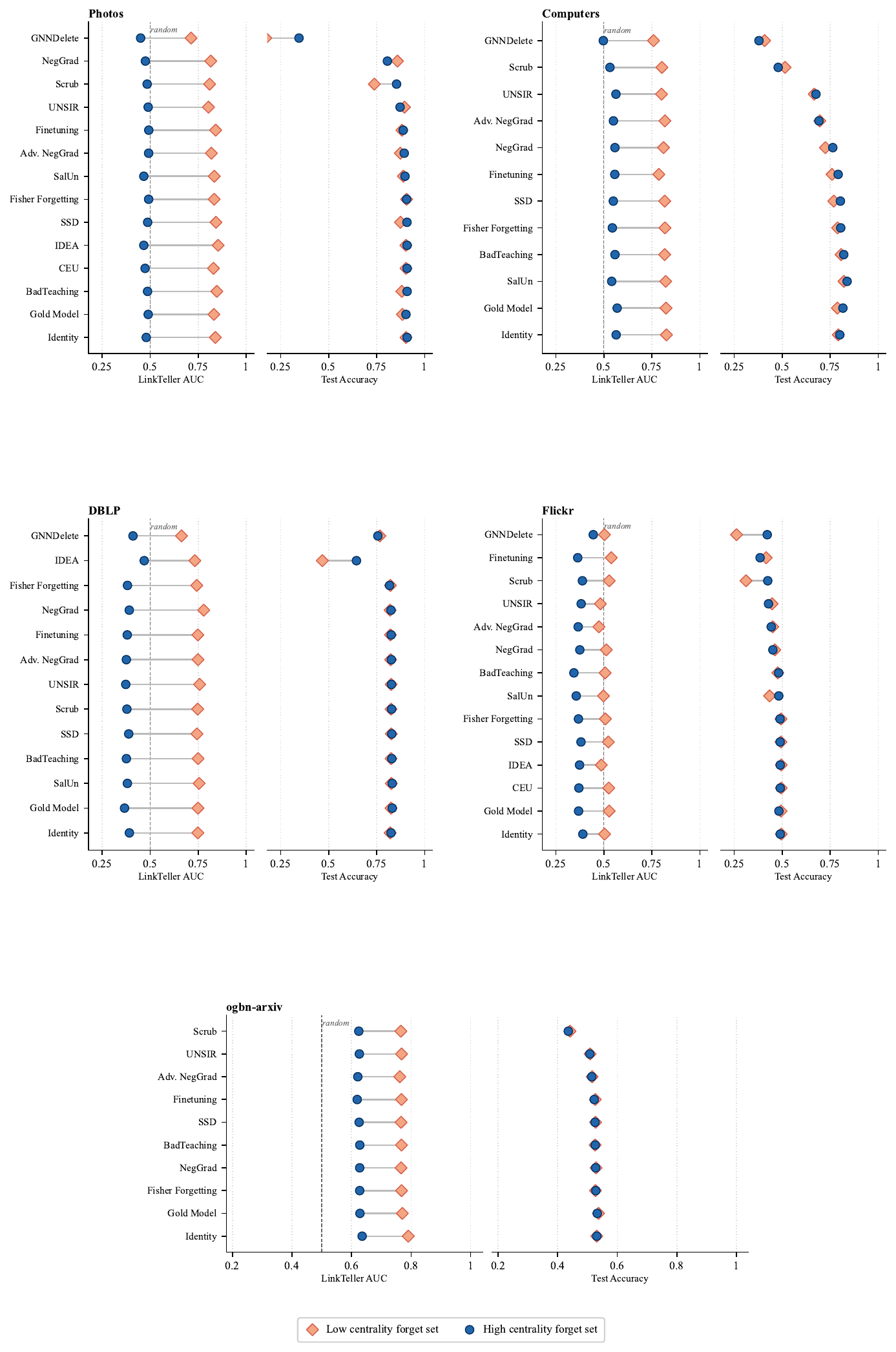}
    \caption{Per-method LinkTeller AUC (left) and test accuracy (right) for all five datasets under GCN, for low-centrality (orange diamonds) and high-centrality (blue circles) forget sets.
    The dashed line marks the random-chance AUC baseline of 0.5.
    GNNDelete is included for completeness but excluded from statistical analysis.}
    \label{fig:rq2_all_datasets}
\end{figure}

Figure~\ref{fig:rq2_all_datasets} expands Figure~3 of the main paper to all five benchmark datasets, showing the individual LinkTeller AUC and test accuracy for every unlearning method under the low- and high-centrality forget set settings.

The low-high centrality gap in LinkTeller AUC is consistent across all datasets and all methods: low-centrality forget sets (orange diamonds) yield higher AUC values, while high-centrality forget sets (blue circles) push most methods toward the random baseline at 0.5.
This is a strong structural result: no method is able to protect low-centrality edges from LinkTeller inference, regardless of the specific unlearning strategy applied.
The right panels confirm that test accuracy is nearly identical between the two settings on every dataset, reinforcing that accuracy cannot distinguish this failure mode.


\subsection{Complementary Architecture Trends}
\label{app:complementary_architectures}

\begin{figure}[t]
    \centering
    \includegraphics[width=\textwidth]{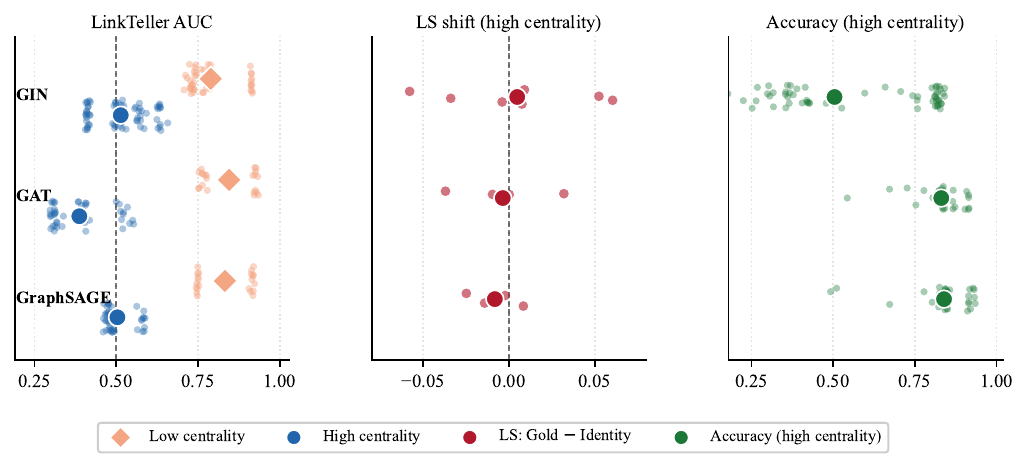}
    \caption{Aggregated complementary results for \textsc{GIN}, \textsc{GAT}, and \textsc{GraphSAGE}.}
    \label{fig:appendix_complementary_architectures}
\end{figure}

To complement the GCN and SGC-centric benchmark view in the main paper, Figure~\ref{fig:appendix_complementary_architectures} show results of
methods for \textsc{GIN}, \textsc{GAT}, and \textsc{GraphSAGE}.

First, the low-/high-centrality bifurcation in LinkTeller remains architecture-agnostic:
LinkTeller is extremely high, while the high-centrality setting collapses to a much lower range for all three architectures.
Second, utility sensitivity is architecture-dependent:
\textsc{GraphSAGE} and \textsc{GAT} retain higher hard-setting accuracy overall, while \textsc{GIN} exhibits a visibly broader spread, reflecting stronger dependence on graph structure.

These results show that the main qualitative conclusions of \our persist beyond GCN to other GNN architectures.

\revision{Finally, Tables \ref{tab:runtime_GCN},\ref{tab:runtime_gin},\ref{tab:runtime_SGC} show, respectively, the running times of all the methods on GCN, GIN, and SGC architectures.  The tables show that, other than the certified methods, two non-certified methods fail to scale: UNSIR and SalUn.}

\begin{table}[t]
\centering
\caption{RunTime comparison on GCN architecture.}
\label{tab:runtime_GCN}
\begin{tabular}{lccccc}
\toprule
Method & DBLP & Photos & Computers & Flickr & ogbn-arxiv \\
\midrule
\textbf{Gold Model -- GCN} & \textbf{9.1} & \textbf{12.4} & \textbf{17.1} & \textbf{14.0} & \textbf{45.6} \\
Finetuning    & 0.8 & 7.9  & 8.9  & 2.2  & 9.5  \\
NegGrad       & 0.8 & 7.2  & 8.9  & 3.8  & 9.6  \\
AdvNegGrad    & 0.9 & 6.9  & 9.0  & 3.7  & 9.9  \\
SCRUB         & 0.9 & 6.5  & 9.2  & 1.8  & 10.1 \\
BadTeaching   & 0.9 & 7.1  & 9.1  & 3.8  & 12.3 \\
SSD           & 0.9 & 6.8  & 9.0  & 1.6  & 9.9  \\
Fisher        & 1.1 & 9.2  & 10.5 & 21.1 & 13.4 \\
UNSIR         & 2.5 & 12.3 & 17.6 & 78.3 & 77.3 \\
SalUn         & 10.1 & 12.7 & 31.2 & 6.2 & 49.9 \\
GNNDelete     & 1.2 & 5.1  & 4.8  & 17.0 & 20.9 \\
IDEA          & 5.6 & 6.6  & 5.8  & 24.0 & 42.0 \\
\bottomrule
\end{tabular}
\end{table}

\begin{table}[t]
\centering
\caption{RunTime comparison on GIN architecture.}
\label{tab:runtime_gin}
\begin{tabular}{lccccc}
\toprule
Method & DBLP & Photos & Computers & Flickr & ogbn-arxiv \\
\midrule
\textbf{Gold Model -- GIN} & \textbf{7.2} & \textbf{13.9} & \textbf{23.6} & \textbf{31.1} & \textbf{100} \\
Finetuning    & 1.1 & 6.5  & 8.0  & 37.6 & 11.7 \\
NegGrad       & 1.1 & 7.0  & 8.2  & 39.3 & 20.4 \\
AdvNegGrad    & 1.5 & 6.6  & 9.0  & 39.0 & 23.9 \\
SCRUB         & 2.3 & 6.6  & 9.7  & 46.6 & 20.6 \\
BadTeaching   & 2.0 & 7.8  & 9.5  & 48.3 & 23.6 \\
SSD           & 1.6 & 6.7  & 8.7  & 39.6 & 24.5 \\
Fisher        & 1.1 & 9.6  & 11.0 & 72.7 & 325 \\
UNSIR         & 8.9 & 11.7 & 23.3 & 139  & 433 \\
SalUn         & 10.9 & 12.7 & 30.3 & 80.7 & 106 \\
IDEA          & 5.6 & 6.0  & 24.0 & 25.2 & 42.0 \\
\bottomrule
\end{tabular}
\end{table}

\begin{table}[t]
\centering
\caption{Runtime comparison on SGC architecture.}
\label{tab:runtime_SGC}
\begin{tabular}{lccccc}
\toprule
Method & DBLP & Photos & Computers & Flickr & ogbn-arxiv \\
\midrule
\textbf{Gold Model -- SGC} & \textbf{18.4} & \textbf{19.7} & \textbf{35.5} & \textbf{56.3} & \textbf{38.5} \\
Finetuning    & 2.1  & 3.9    & 11.8  & 17.9   & 11.5 \\
NegGrad       & 2.2  & 4.4    & 9.6   & 19.0   & 11.8 \\
AdvNegGrad    & 2.3  & 4.3    & 9.9   & 17.3   & 12.1 \\
SCRUB         & 2.3  & 4.2    & 9.2   & 18.7   & 12.5 \\
BadTeaching   & 2.5  & 4.5    & 9.8   & 18.6   & 15.0 \\
SSD           & 2.2  & 4.3    & 10.0  & 18.1   & 12.2 \\
Fisher        & 2.4  & 5.2    & 10.6  & 19.0   & 17.0 \\
UNSIR         & 5.6  & 8.1    & 18.1  & 90.2   & 83.0 \\
SalUn         & 4.0  & 7.1    & 16.4  & 34.3   & 51.0 \\
IDEA          & 4.2  & 5.3    & 11.8  & 22.5   & 49.9 \\
CEU           & 30.8 & 22.9   & 69.6  & 78.0   & 64.8 \\
CGU           & 35.5 & 316.0  & 247.0 & 1576.0 & 59.3 \\
ScaleGUN      & 51.8 & 3096.0 & 162.0 & 3008.0 & 71.5 \\
\bottomrule
\end{tabular}
\end{table}


\clearpage
\newpage
\section*{NeurIPS Paper Checklist}

\begin{enumerate}

\item {\bf Claims}
    \item[] Question: Do the main claims made in the abstract and introduction accurately reflect the paper's contributions and scope?
    \item[] Answer: \answerYes{} 
    \item[] Justification: The main claims are made clear in the abstract and introduction and are followed through in later sections. All claims match experimental results.

\item {\bf Limitations}
    \item[] Question: Does the paper discuss the limitations of the work performed by the authors?
    \item[] Answer: \answerYes{} 
    \item[] Justification: We make our positioning with respect to other benchmarks clear in the paper, and we discuss limitations in the Conclusions. 

\item {\bf Theory assumptions and proofs}
    \item[] Question: For each theoretical result, does the paper provide the full set of assumptions and a complete (and correct) proof?
    \item[] Answer: \answerNA{} 
    \item[] Justification: This paper is a benchmark paper with empirical evaluations. No theoretical results are presented. However, we back empirical claims with large-scale evaluation and statistical testing.

    \item {\bf Experimental result reproducibility}
    \item[] Question: Does the paper fully disclose all the information needed to reproduce the main experimental results of the paper to the extent that it affects the main claims and/or conclusions of the paper (regardless of whether the code and data are provided or not)?
    \item[] Answer: \answerYes{} 
    \item[] Justification: The code, along with instructions to run it, is made publicly available and the link is given. We discuss our settings, both hardware and Software, in Section 5.

\item {\bf Open access to data and code}
    \item[] Question: Does the paper provide open access to the data and code, with sufficient instructions to faithfully reproduce the main experimental results, as described in supplemental material?
    \item[] Answer: \answerYes{} 
    \item[] Justification: The code repository is made available and instructions are available in the repository.

\item {\bf Experimental setting/details}
    \item[] Question: Does the paper specify all the training and test details (e.g., data splits, hyperparameters, how they were chosen, type of optimizer) necessary to understand the results?
    \item[] Answer: \answerYes{} 
    \item[] Justification: We detail all the necessary details in Section 5, including data splits, hyperparameters, and optimizer.

\item {\bf Experiment statistical significance}
    \item[] Question: Does the paper report error bars suitably and correctly defined or other appropriate information about the statistical significance of the experiments?
    \item[] Answer: \answerYes{} 
    \item[] Justification: We report statistical tests for all results in Section \ref{sec:results}. 

\item {\bf Experiments compute resources}
    \item[] Question: For each experiment, does the paper provide sufficient information on the computer resources (type of compute workers, memory, time of execution) needed to reproduce the experiments?
    \item[] Answer: \answerYes{} 
    \item[] Justification: All experiments were run on the same hardware that we describe in detail. Runtime is an explicit part of experimentation.
    
\item {\bf Code of ethics}
    \item[] Question: Does the research conducted in the paper conform, in every respect, with the NeurIPS Code of Ethics \url{https://neurips.cc/public/EthicsGuidelines}?
    \item[] Answer: \answerYes{} 
    \item[] Justification: We read the Ethics Guidelines and we made sure to comply in every respect.

\item {\bf Broader impacts}
    \item[] Question: Does the paper discuss both potential positive societal impacts and negative societal impacts of the work performed?
    \item[] Answer: \answerYes{} 
    \item[] Justification: The paper discusses positive societal impacts in supporting privacy regulations such as GDPR  (Sections 1 and 5). The paper also discusses negative societal impacts and security considerations by studying link inference attacks, adversarial reconstruction of forgotten edges, which is the main motivation of the benchmark.
    
\item {\bf Safeguards}
    \item[] Question: Does the paper describe safeguards that have been put in place for responsible release of data or models that have a high risk for misuse (e.g., pre-trained language models, image generators, or scraped datasets)?
    \item[] Answer: \answerNA{} 
    \item[] Justification: We do not release new data nor models.

\item {\bf Licenses for existing assets}
    \item[] Question: Are the creators or original owners of assets (e.g., code, data, models), used in the paper, properly credited and are the license and terms of use explicitly mentioned and properly respected?
    \item[] Answer: \answerYes{} 
    \item[] Justification: We cite and reference all the authors of the original methods used in the benchmark. Their methods were faithfully implemented.

\item {\bf New assets}
    \item[] Question: Are new assets introduced in the paper well documented and is the documentation provided alongside the assets?
    \item[] Answer: \answerYes{} 
    \item[] Justification: We release the benchmark code, which is open and well-documented.

\item {\bf Crowdsourcing and research with human subjects}
    \item[] Question: For crowdsourcing experiments and research with human subjects, does the paper include the full text of instructions given to participants and screenshots, if applicable, as well as details about compensation (if any)? 
    \item[] Answer: \answerNA{} 
    \item[] Justification: There was no crowdsourcing involved in this benchmark study.

\item {\bf Institutional review board (IRB) approvals or equivalent for research with human subjects}
    \item[] Question: Does the paper describe potential risks incurred by study participants, whether such risks were disclosed to the subjects, and whether Institutional Review Board (IRB) approvals (or an equivalent approval/review based on the requirements of your country or institution) were obtained?
    \item[] Answer: \answerNA{} 
    \item[] Justification: There was no crowdsourcing involved in this benchmark study.

\item {\bf Declaration of LLM usage}
    \item[] Question: Does the paper describe the usage of LLMs if it is an important, original, or non-standard component of the core methods in this research? Note that if the LLM is used only for writing, editing, or formatting purposes and does \emph{not} impact the core methodology, scientific rigor, or originality of the research, declaration is not required.
    \item[] Answer: \answerYes{} 
    \item[] Justification: A dedicated section in the appendix specifies the usage of LLMs in the preparation of this work.

\end{enumerate}

\end{document}